\documentclass[pdflatex,sn-mathphys-num]{sn-jnl}

\usepackage{graphicx}%
\usepackage{multirow}%
\usepackage{amsmath,amssymb,amsfonts}%
\usepackage{amsthm}%
\usepackage{mathrsfs}%
\usepackage[title]{appendix}%
\usepackage{xcolor}%
\usepackage{textcomp}%
\usepackage{manyfoot}%
\usepackage{booktabs}%
\usepackage{algorithm}%
\usepackage{algorithmicx}%
\usepackage{algpseudocode}%
\usepackage{listings}%
\usepackage{makecell}%
\usepackage{booktabs}%
\usepackage{multirow}%
\usepackage{placeins}%
\usepackage{lmodern}

\begin{document}

\title[Article Title]{Natural Language Processing for Cardiology: A Narrative Review}

\author*[1]{\fnm{Kailai} \sur{Yang}}\equalcont{These authors contributed equally to this work.}\email{kailai.yang@manchester.ac.uk}

\author[1]{\fnm{Yan} \sur{Leng}}\equalcont{These authors contributed equally to this work.}

\author[1]{\fnm{Xin} \sur{Zhang}}

\author[1]{\fnm{Tianlin} \sur{Zhang}}

\author[1]{\fnm{Paul} \sur{Thompson}}

\author[2,3]{\fnm{Bernard} \sur{Keavney}}

\author[2,3]{\fnm{Maciej} \sur{Tomaszewski}}

\author[1]{\fnm{Sophia} \sur{Ananiadou}}

\affil[1]{\orgdiv{Department of Computer Science}, \orgname{The University of Manchester}, \orgaddress{\street{National Centre for Text Mining}, \city{Manchester}, \country{UK}}}

\affil[2]{\orgdiv{Division of Cardiovascular Sciences, Faculty of Biology, Medicine and Health}, \orgname{The University of Manchester}, \orgaddress{ \city{Manchester}, \country{UK}}}

\affil[3]{\orgdiv{Manchester University NHS Foundation Trust}, \orgname{Manchester Royal Infirmary}, \city{Manchester}, \country{UK}}


\abstract{Cardiovascular diseases are becoming increasingly prevalent in modern society, with a profound impact on global health and well-being. These Cardiovascular disorders are complex and multifactorial, influenced by genetic predispositions, lifestyle choices, and diverse socioeconomic and clinical factors. Information about these interrelated factors is dispersed across multiple types of textual data, including patient narratives, medical records, and scientific literature. Natural language processing (NLP) has emerged as a powerful approach for analysing such unstructured data, enabling healthcare professionals and researchers to gain deeper insights that may transform the diagnosis, treatment, and prevention of cardiac disorders. This review provides a comprehensive overview of NLP research in cardiology from 2014 to 2025. We systematically searched six literature databases for studies describing NLP applications across a range of cardiovascular diseases. After a rigorous screening process, we identified 265 relevant articles. Each study was analysed across multiple dimensions, including NLP paradigms, cardiology-related tasks, disease types, and data sources. Our findings reveal substantial diversity within these dimensions, reflecting the breadth and evolution of NLP research in cardiology. A temporal analysis further highlights methodological trends, showing a progression from rule-based systems to large language models. Finally, we discuss key challenges and future directions, such as developing interpretable LLMs and integrating multimodal data. To the best of our knowledge, this review represents the most comprehensive synthesis of NLP research in cardiology to date.}




\maketitle

\section{Introduction}\label{sec1}

Cardiovascular diseases (CVDs) comprise a group of disorders that affect the heart and blood vessels, posing a major global health challenge. The most common clinical manifestations include coronary artery disease, heart failure, arrhythmias, valvular heart disease, and hypertension. These conditions can significantly diminish an individual’s quality of life and overall well-being, with their prevalence continuing to rise due to aging populations and lifestyle changes \citep{roth2020global}. According to the World Health Organization, CVDs are the leading cause of death worldwide, accounting for an estimated 17.9 million deaths each year \citep{who2021cvd}. Consequently, early detection and effective management of CVDs are essential to improving patient outcomes and mitigating their burden on healthcare systems.

There are many types of text data that contain valuable information about the cardiovascular system. These include electronic health records (EHRs), clinical notes, radiology reports, research papers, and patient-reported outcomes,etc. In recent years, natural language processing (NLP), a branch of artificial intelligence (AI), has emerged as a powerful tool for analyzing and managing large-scale textual data in the field of cardiology. With the help of NLP algorithms, healthcare professionals can gain valuable insights from vast amounts of unstructured data, thus providing the potential to revolutionize approaches to cardiac diagnosis, treatment, and prevention. NLP techniques have been successfully applied to various tasks in cardiology, including the extraction of important cardiac measurements from echocardiography reports \citep{nath2016natural}, classifying the severity of coronary artery disease from clinical narratives \citep{gao2019automated},  predicting heart failure readmissions based on discharge summaries \citep{golas2018machine}, identifying risk factors for CVDs, and supporting clinical decision-making \citep{chen2020natural}. The rapidly evolving nature of NLP research promises to contribute towards further enhancements to patient care, advances in clinical research, and the development of more personalized and effective cardiovascular interventions \citep{ahmed2025primer,ferreira2025applications}.

The active nature of NLP research within the cardiology domain has already given rise to some brief reviews and surveys. For example, \citet{turchioe2022systematic} focused on identifying articles that describe the application of NLP techniques to EHRs referring to cardiology issues during the period 2015-2020. The review revealed that rule-based methods were dominant, although a smaller number of approaches based on traditional machine learning were also identified. The rapidly evolving nature of the NLP field has led to more reviews on the emerging technique of large language models (LLMs)~\citep{gala2023utility,ferreira2025applications,boonstra2024artificial} within the cardiology domain, which have rapidly grown to be a dominant paradigm of many NLP approaches. Their training process on vast amounts of textual data facilitates sophisticated language understanding and text generation capabilities, which have been successfully applied to carry out a wide variety of related tasks.

Despite the utility of the above reviews, there is currently a lack of a survey that provides a comprehensive account of the evolution of NLP within the cardiology domain that examines the full range of NLP techniques that have been applied in the field of cardiology over an extended time period, and how these techniques have been applied to address a variety of cardiology-related applications concerning a broad set of cardiovascular diseases. In response, this review article provides a detailed overview of the NLP applications within the cardiology domain over the last decade. In contrast to \citet{turchioe2022systematic}, which only identified 37 NLP-related articles containing general terms relating to cardiology (i.e., \textit{cardiology}, \textit{cardiac} or \textit{cardiovascular}), we have conducted a extensive search that covers a range of more specific cardiovascular diseases. This allows us to consider the impact of NLP in greater depth across various branches of cardiology. Specifically, we cover the period 2014 - 2025, which overlaps with, but extend beyond, the time span covered by \citet{turchioe2022systematic}, leading to coverage of crucial emerging trends such as LLM-based methods and novel perspectives such as retrieval-augmented generation. Our extended search parameters have also resulted in the identification of a much larger 265 number of relevant studies compared to previous reviews, thus permitting an in-depth analysis of the impact of NLP across a range of cardiovascular sub-fields, such as heart failure \citep{girouard2025clinical} and ECG diagnosis \citep{ansari2025survey}. The wide span of related works that we cover also allows us to comprehensively analyze the evolution of NLP methods in cardiology, providing a systematic landscape of advancing NLP techniques -- from traditional machine learning to LLMs -- and their influence on various cardiovascular sub-fields.


In summary, this review is aimed at answering the following research questions:
\begin{itemize}
 
\item What types of NLP paradigms have been employed to address various text-related tasks and applications in cardiology?

\item To what extent have NLP approaches been applied across different cardiovascular diseases and types of textual sources?  

\item What is the evolving landscape and emerging trends of NLP approaches for cardiology?

\item What is the key challenges and most promising directions for future research for more effective NLP applications in cardiology?
\end{itemize}

\section{Search Methodology}\label{sec2}

\subsection{Search strategy}

In order to identify relevant articles, we conducted a comprehensive search using six literature search platforms (i.e,  PubMed, Scopus, Web of Science, IEEE Xplore, ACM Digital Library, and DBLP computer science bibliography) to find articles written in English and published between July 2014 and October 2025. 

We formulated a search query based on two sets of keywords, one of which includes words and phrases relating to a range of different cardiovascular diseases. In contrast, the other set consists of words and phrases typically found in articles discussing NLP methods, including those that employ state-of-the-art LLMs. The query aimed to retrieve articles containing at least one word or phrase from each of these two sets since such articles are likely to describe the application of NLP methods within the context of cardiovascular studies. Accordingly, the search query was formulated by conjoining each of the keywords/phrases in the sets using the Boolean operator “OR" and combining the two sets of keywords using the Boolean operator “AND". The two sets of keywords are shown in Table~\ref{tab:keywords}. Among the 17 terms related to cardiovascular disease are general terms for disease in cardiology (e.g., heart disease ), common specific cardiovascular diseases (e.g., congenital heart defect, coronary artery disease, heart failure), and names of various radiology tests (e.g., electrocardiogram, echocardiography,  cardiac magnetic resonance). We conducted the searches until October 2025.

\begin{table}[ht]
\caption{The keywords for literature search.}\label{tab:keywords}
\begin{tabular}{@{}p{0.2\textwidth}p{0.8\textwidth}@{}}
\toprule
Category & Keywords \\
\midrule
Cardiovascular disease (1) & cardiac, cardiology, cardiovascular, heart disease \\
 & congenital heart defect, coronary artery disease, heart failure, atrial fibrillation, arrhythmias \\
 & electrocardiogram, ECG, echocardiography, echocardiogram, cardiography \\
 & heart magnetic resonance, cardiac magnetic resonance, cardiovascular magnetic resonance \\
\midrule
Methods (2) & Natural language processing, NLP, large language models, large language model, LLMs, text mining, text analysis \\
\midrule
Search query & (1) AND (2) \\
\botrule
\end{tabular}
\end{table}

\subsection{Filtering strategy}
\begin{figure}[ht]
  \centering \includegraphics[width=8cm]{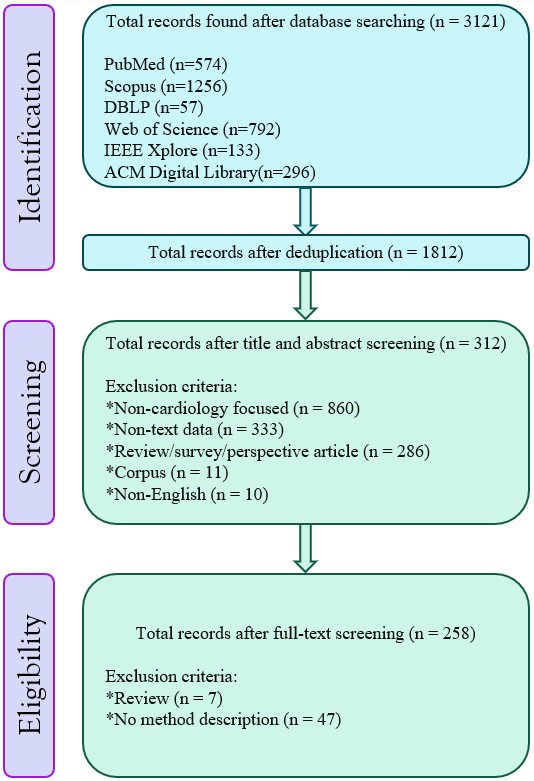}
  \caption{A three-stage pipeline of the article selection process.}
  \label{fig:selection}
\end{figure}

Figure \ref{fig:selection} provides an overview of our filtering strategy. A total of 3121 articles were retrieved from the six databases, of which 1812 records were retained after deduplication. We subsequently employed the NLP-driven RobotAnalyst tool \citep{przybyla2018prioritising} to screen the titles and abstracts of the remaining articles. RobotAnalyst is a tool that minimizes the human workload in the screening phase of reviews by prioritizing the most relevant articles for cardiovascular disease based on relevancy feedback and active learning \citep{o2015using, miwa2014reducing}.

The goal of the screening process was to remove records whose abstract made it clear that the article met one of the following exclusion criteria: (1) the focus was not on cardiology;
(2) a method making use of waveform, sound or image data, rather than textual data, was described; (3) the article constituted a review, survey or perspective article; (4) the article described a corpus without any new method; (5) the full text was not available in English.

Following the title/abstract screening process, 312 records remained, whose full text was subsequently reviewed. This final review aimed to ensure that only articles describing the application of specific NLP methods to unstructured, cardiology-related textual data were retained for detailed analysis. Articles that did not describe such methods, or which constituted review, survey or perspective papers, were excluded. 

The full-text screening process resulted in the exclusion of a further 54 articles. The remaining 258 articles formed the basis for the detailed survey that is described in the rest of this article.

\section{Findings}\label{sec3}

\begin{figure*}[htpb]
\centering
\includegraphics[width=12cm]{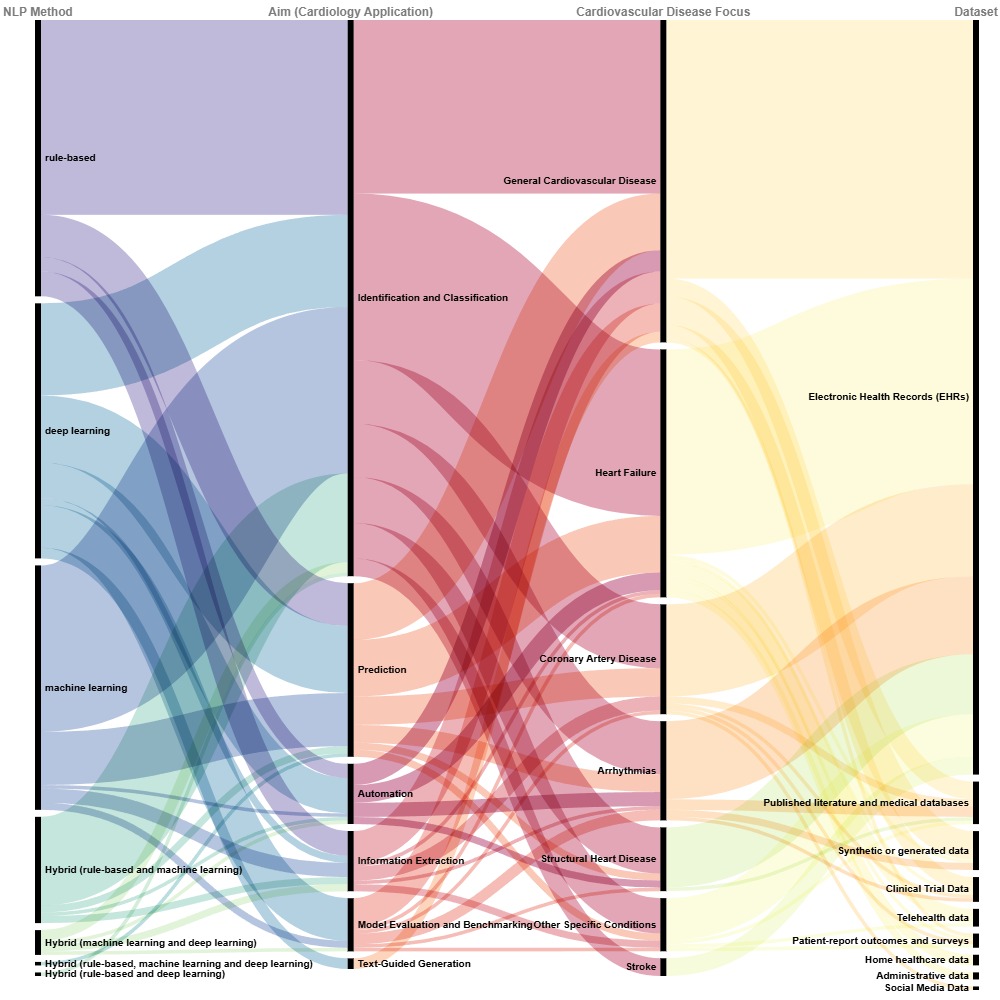}
\caption{A general overview of the diversity of research reported.}
\label{fig:sankey}
\end{figure*}

The Sankey diagram in Figure \ref{fig:sankey} provides a general overview of the diversity of research reported across the 258 articles that we have reviewed. The diagram is intended to allow major research trends to be identified, and for the richness and complexity of NLP-based cardiology research to be fully appreciated. 

We categorized each article according to four different aspects/dimensions of the research reported, i.e.,
\begin{itemize}
    \item The type of NLP methods employed;
    \item The cardiology-related tasks/applications addressed by the research;
    \item The type of cardiovascular diseases that forms the focus of the research;
    \item The type of datasets to which the reported NLP method was applied.
\end{itemize}

In Figure \ref{fig:sankey}, each of these dimensions is represented as a vertical bar, which is divided into sections, each corresponding to a specific value for the dimension. The length of each section of the bar indicates the number of articles that are categorized according to the corresponding value.   

The streamlines that flow between the vertical bars illustrate pairwise relationships that hold between values of the different dimensions studied, i.e:

\begin{itemize}
    \item Which NLP methods/techniques have been used to address different types of cardiology-related tasks/applications;
    \item Which types of cardiovascular diseases have formed the focus of each type of task/application;
    \item Which types of text have been processed in the context of each type of disease.
\end{itemize}

The width of each stream indicates the number of articles that provide evidence of the relationship between the connected values.  

In most cases, multiple streams originate from each value in a given source dimension and flow to many different values in the connected target dimension. Conversely, multiple streams usually converge on each value in the target dimension. These many-to-many connections serve to illustrate the richness and complex nature of NLP research efforts within the cardiology domain. The figure clearly illustrates that a wide range of cardiology-related tasks/applications have been addressed by different studies and in most cases, multiple types of NLP methods and techniques have been investigated as a means to solve each of these tasks. Each task/application has generally been considered in the context of multiple cardiovascular diseases, while different studies have focused on processing information related to each of these diseases in diverse types of textual sources.

\subsection{Data Sources}

Important information relating to cardiovascular diseases can occur in a wide range of different types of text. Generally, each different type of text possesses specific characteristics and features and brings its own challenges for effective automated processing by NLP methods. Accordingly, it is desirable for NLP techniques that are designed to address specific tasks and/or diseases to be trained and evaluated using a diverse set of different text types. This can help to ensure the robustness of methods and to minimise the potential overlook of vital information. We visualised the diverse types of data used in the 258 papers in this section to help understand the mainstream and relatively less-adopted data types in NLP applications for cardiovascular diseases.

\begin{figure}[ht]
  \centering
  \includegraphics[width=9cm]{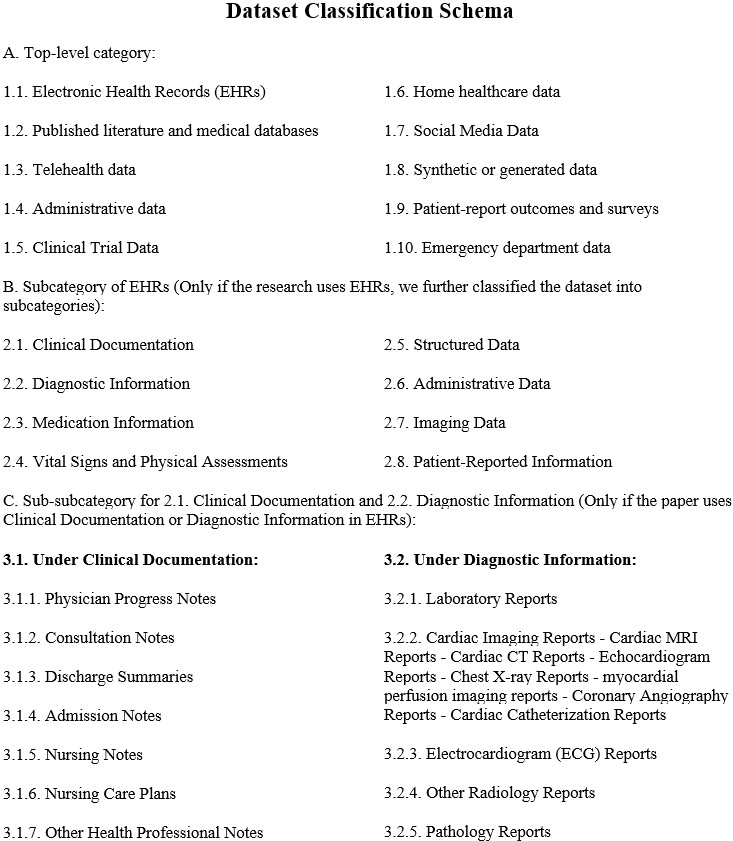}
  \caption{The hierarchical dataset classification schema for classification of data sources.}
  \label{fig:Dataset_Classification}
\end{figure}

We proposed a three-level hierarchy shown in Figure \ref{fig:Dataset_Classification}, which allows for a detailed and flexible classification of data sources, particularly for EHRs and clinical notes/reports. It's grounded in the structure provided by Häyrinen et al.  \cite{hayrinen2008definition}, while also accommodating the specific needs of cardiology-focused research. When classifying a data source, we first identify whether it's from EHRs or another category, then specify the subcategory within EHRs, and finally, if it's clinical documentation or diagnostic information, we further classify it into the appropriate sub-subcategory. This approach provides a clear, evidence-based method for classifying datasets in this review, with a particular focus on cardiology-related data sources within EHRs. The proposed classification schema is proposed after a preview of all 258 papers and is adjusted to fit all the papers.

\subsubsection{Top-level Category}

\begin{figure}[ht]
  \centering
  \includegraphics[width=\textwidth]{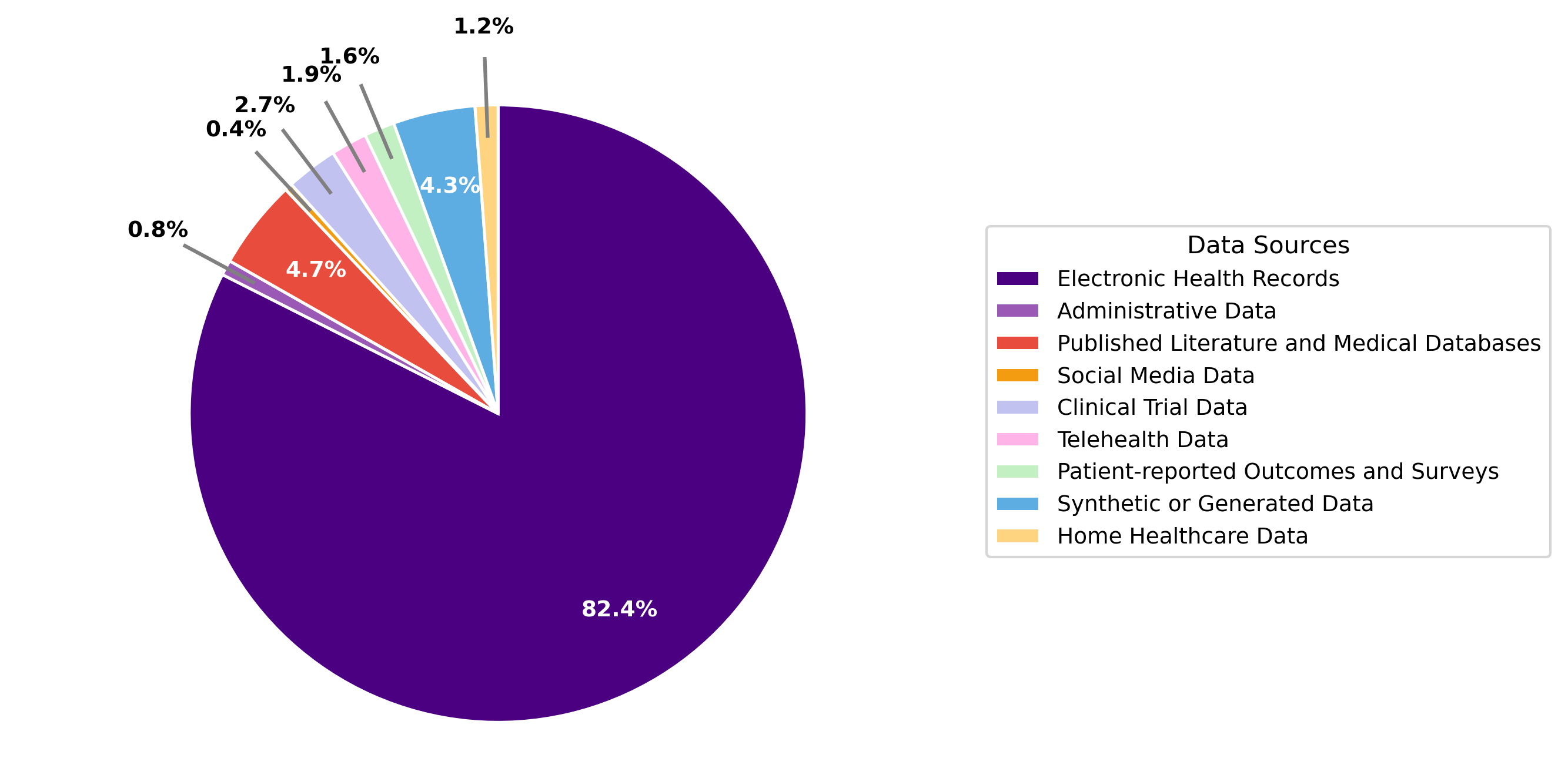}
  \caption{Pie chart of the distribution of data source categories in Cardiology NLP research.}
  \label{fig:tree}
\end{figure}

Figure \ref{fig:tree} illustrates the range and distribution of the different types of cardiology text to which NLP methods have been applied in the articles that we have reviewed. The classification was done manually.

As shown in figure \ref{fig:tree}, a total of 9 different types of textual data have been considered in the reviewed articles. The vast majority of studies make use of Electronic Health Records (EHRs), which account for 82.4\% of the data sources used in the studies included in the review. Published literature and medical databases constitute the second largest category at 4.7\%, followed closely by synthetic or generated data at 4.3\%. Clinical Trial Data represents 2.7\% of the sources, while telehealth data accounts for 1.9\%. Patient-reported outcomes and surveys make up 1.6\%, and home healthcare data represents 1.2\%. Administrative data (0.8\%) and social media data (0.4\%) represent smaller portions of the data sources used.

This distribution shows that while EHRs heavily dominate as the primary data source, other types of textual data have been considered, but only by a relatively small number of studies. For example, while social media represents a potentially vast source of information, it remains largely unexplored, accounting for only 0.4\% of the studies reviewed.

\paragraph{Electronic Health Records (EHRs)}
Electronic Health Records (EHRs) are comprehensive digital records of patients' medical histories, diagnoses, medications, treatment plans, and other health-related information \cite{shickel2017deep}. EHRs have become a valuable resource for NLP research in cardiology, due to their rich longitudinal data and potential for large-scale analysis \citep{doi:10.1016/j.jchf.2023.02.012}.
For example, \citet{RN2214} developed EchoInfer, an NLP system that accurately extracted 80 cardiac concepts from 15,116 echocardiogram reports at a single academic medical centre, with limited testing on additional datasets from the same institution. \citet{RN2215} applied NLP techniques to EHR data from 118,002 patients to study trends in hospitalizations for worsening heart failure, which illustrates the potential scalability of NLP approaches to large EHR datasets. \citet{brennan2025large} successfully applied LLMs to detect atrial fibrillation progression in electronic clinical notes at mass scale.

\paragraph{Published Literature and Medical Databases}
Published literature and medical databases are rich sources of curated medical knowledge and research findings, which are particularly useful as the basis for developing NLP systems that can understand and interpret complex medical concepts and relationships \citep{henry2020building}. An example of a study that utilizes published literature is the study by \citet{RN2363}, which used 120 cardiology questions from MKSAP-19 to evaluate the clinical decision-making ability of LLMs. This approach demonstrates how NLP can be applied to assess and potentially enhance medical education and decision support in cardiology.

\paragraph{Synthetic or Generated Data} Synthetic or generated data refers to artificially created healthcare data that mimics the statistical properties and patterns of real patient data while maintaining privacy protections. This type of data is becoming increasingly important for developing and testing healthcare AI systems, particularly when real patient data is scarce or privacy concerns limit data sharing. Synthetic data generation can help address data scarcity and privacy concerns in healthcare research. For instance, \citet{RN2517} uses synthetic data for automated extraction of family history information from Norwegian clinical text. \citet{sowa2025fine} utilized LLM-modified synthetic data for fine-tuning EchoGPT, an effective LLM model that can generate high-quality echocardiography reports.

\paragraph{Clinical Trial Data} 
Clinical trial data represents highly structured and carefully collected information from controlled research studies investigating medical interventions, treatments, or diagnostic approaches \citep{zarin2016trial,RN2517}. This data type is characterized by its rigorous collection protocols, standardized formats, and detailed documentation of patient outcomes and adverse events. Clinical trial data is considered the gold standard for evidence-based medicine, though its generalizability may be limited by strict inclusion/exclusion criteria. For example, Cunningham et al. \cite{RN5} utilized the INVESTED trial data \citep{vardeny2018high} to validate and fine-tune a Natural Language Processing (NLP) model for automatically identifying heart failure hospitalizations.

\paragraph{Telehealth Data} 
Telehealth data represents clinical information captured during remote healthcare encounters, including virtual consultations, remote monitoring, and digital health interactions between patients and healthcare providers \citep{doraiswamy2020use}. This data type encompasses both structured measurements (like vital signs and device readings) and unstructured content (such as provider notes and patient messages), offering unique insights into healthcare delivery and patient management outside traditional clinical settings. The increasing adoption of telehealth, particularly accelerated by the COVID-19 pandemic, has made this data source increasingly valuable for understanding healthcare delivery patterns, patient engagement, and clinical outcomes. For example, \citet{RN2449} developed classifiers for automated categorization of medication-related notes in heart failure telehealth patients.

\paragraph{Patient-reported Outcomes and Surveys} 
Patient-reported outcomes and surveys capture direct feedback from patients about their health status, symptoms, quality of life, and healthcare experiences \citep{basch2017patient}. This data type provides crucial insights into the patient's perspective and subjective experiences that may not be captured in clinical documentation. Patient-reported data is essential for understanding treatment effectiveness and patient satisfaction from the patient's perspective. For instance, Chen et al. \cite{RN2462} combined patient survey data from TRACE-CORE \citep{waring2012transitions} with EHR data to study how post-discharge pain relates to hospital readmission while considering both patient-reported symptoms and documented clinical care.

\paragraph{Home Healthcare Data} 
Home healthcare data encompasses clinical information collected during healthcare services provided in patients' homes, including vital signs, symptoms, medication adherence, and functional assessments \cite{topaz2019mining}. This data type offers unique insights into patients' daily living conditions and health management behaviours outside traditional clinical settings. Home healthcare data is particularly valuable for understanding chronic disease management and patient self-care capabilities. For example, \citet{RN2401} used 2.3 million home healthcare clinical notes to identify patients in home healthcare with heart failure symptoms and poor self-management.

\paragraph{Administrative data} 
Administrative data encompasses structured information generated during healthcare business operations, including insurance claims, billing records, and operational metrics, which is typically maintained in systems separate from but complementary to EHRs \citep{kohane2011using,RN2250}. While primarily collected for billing, reimbursement, and administrative purposes, source allocation, and system-level performance indicators across different healthcare settings, administrative data has become a valuable resource for healthcare research due to its comprehensiveness and longitudinal nature. In the reviewed papers, \citet{RN162} utilised administrative health data to predict the risk of 30-day readmissions in patients with heart failure.

\paragraph{Social Media Data} 
Social media data represents user-generated content from various online platforms where patients and healthcare providers share experiences, symptoms, and health-related information \citep{sinnenberg2017twitter}. This data source provides real-time, unfiltered insights into patient experiences, concerns, and health behaviours outside clinical settings. Social media data analysis can reveal emerging health trends and patient perspectives that might not be captured in traditional clinical documentation. One of our reviewed papers, \citet{RN2325}, used a semi-supervised natural language processing model and unsupervised machine learning techniques to
analyze 5,606 coronary artery calcium-related discussions on Reddit.


\begin{figure}[ht]
  \centering
  \includegraphics[width=\textwidth]{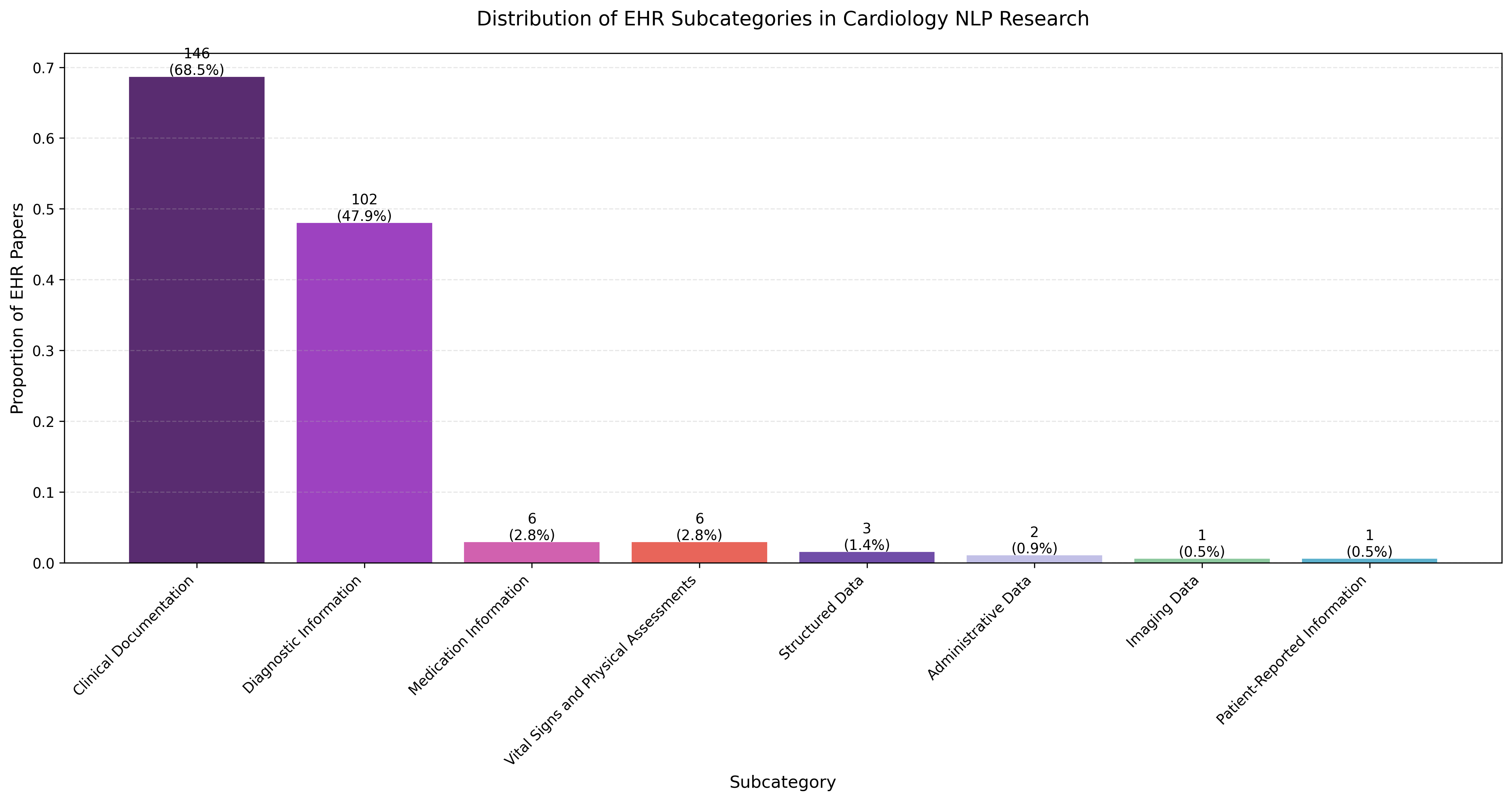}
  \caption{Distribution of different subcategories of EHRs.}
  \label{fig:bar_chart_sub}
\end{figure}

\subsubsection{Subcategories of EHRs}
As there are more than 80\% reviewed papers which focus on Electronic Health Records (EHRs), we further identified eight distinct subcategories within EHR-based studies. Unlike the classification of top categories, a paper could use multiple parts of EHRs. Therefore we conduct multi-class classification here. As shown in Figure \ref{fig:bar_chart_sub}, among the 258 reviewed papers, Clinical Documentation represented the predominant data source (68.5\%, n=146), followed by Diagnostic Information which accounted for nearly half of the papers (47.9\%, n=102). Notably, both Medication Information and Vital Signs and Physical Assessments appeared less frequently, each representing 2.8\% (n=6) of the papers. The remaining data sources were utilized less commonly, with Structured Data appearing in 1.4\% (n=3) of papers, while both Imaging Data and Patient-Reported Information were found in just 0.5\% (n=1) of the papers each.

The distribution pattern reveals that Clinical Documentation dominates the research landscape, being utilized in over two-thirds of EHR-based studies, while Diagnostic Information represents the second most significant data source, appearing in nearly half of the studies. Notably, there exists a striking gap between these top two categories and the remaining six, with all other categories being used in less than 3\% of studies. Other categories are also crucial parts of EHRs and are widely studied using computation methods. However, they are rarely used in our reviewed papers because we excluded papers that didn't use text information, which means papers focused only on structured data or imaging data are not analysed in this research. 

\subsubsection{Clinical Documentation and Diagnostic Information}

\begin{figure}[ht]
  \centering
  \includegraphics[width=\textwidth]{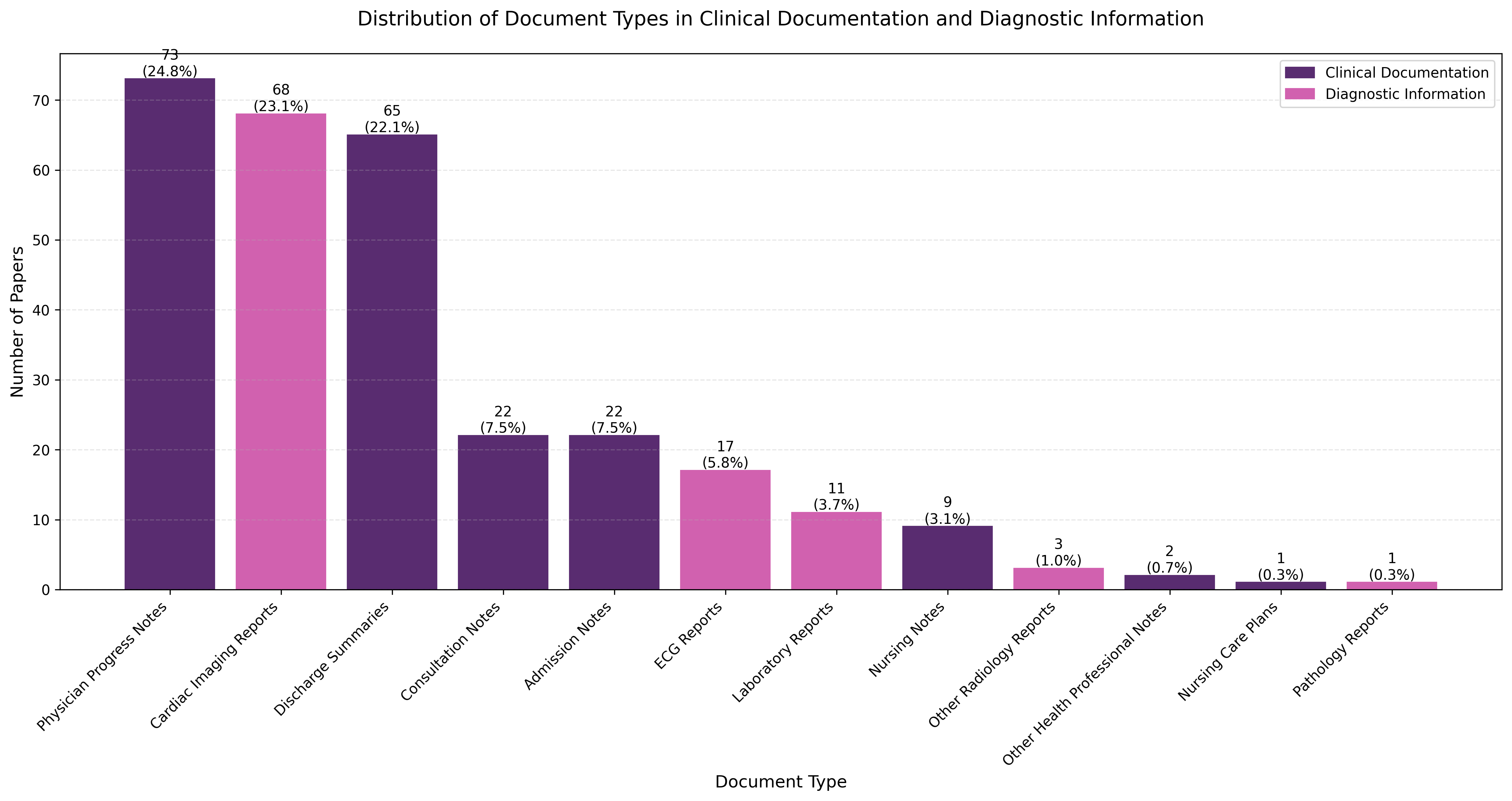}
  \caption{Distribution of sub-categories in clinical documentations and diagnostic information.}
  \label{fig:bar_chart_subsub}
\end{figure}

As Clinical Documentation and Diagnostic Information are the crucial data sources which are widely used in NLP research for Cardiovascular disease, we further classified the two categories into specific sub-subcategories to provide a more granular understanding of the data sources utilized in cardiovascular NLP research. The distributions of different sub-subcategories are shown in Figure \ref{fig:bar_chart_subsub}.

Among Clinical Documentation types, which encompass various notes and summaries created by healthcare providers during patient care, Physician Progress Notes represents the most frequently used document type (24.8\%, n=73), followed by Discharge Summaries (22.1\%, n=65), Consultation Notes and Admission Notes each account for 7.5\% (n=22) of the documents. As for Diagnostic Information types, which consist of various medical test results and imaging interpretations, Cardiac Imaging Reports dominates with 23.1\% (n=68) of all document types, followed by ECG Reports (5.8\%, n=17) and Laboratory Reports (3.7\%, n=11). Notably, there is a substantial drop in frequency after the top three document types, with the remaining types such as Nursing Notes (3.1\%, n=9), Other Radiology Reports (1.0\%, n=3), Other Health Professional Notes (0.7\%, n=2), and both Nursing Care Plans and Pathology Reports (0.3\%, n=1 each) being used much less frequently in the literature.

\subsubsection{Imaging and Radiology Reports}

\begin{figure}[ht]
  \centering
  \includegraphics[width=\textwidth]{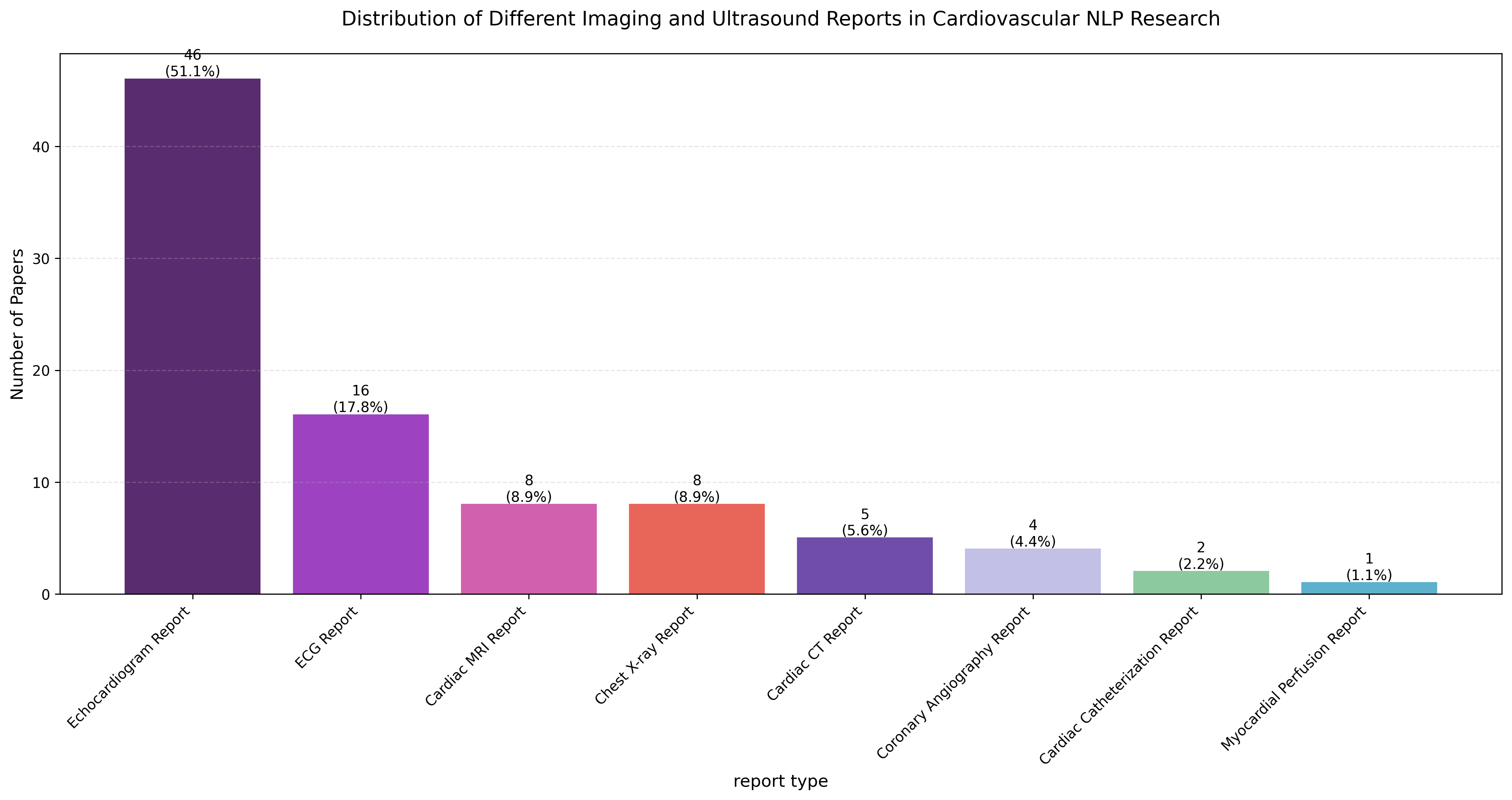}
  \caption{Distribution of sub-categories in cardiac imaging and radiology reports.}
  \label{fig:bar_chart_image}
\end{figure}

Cardiac imaging reports provide detailed information about cardiac structure and function. NLP techniques can help extract and analyze this information efficiently \cite{pons2016natural}. For example, Carthey et al. \cite{RN2456} developed an NLP system to extract various types of phenotypic information from narrative cardiac magnetic resonance (CMR) reports to aid with the diagnosis of hypertrophic cardiomyopathy (HCM). 

Figure \ref{fig:bar_chart_image} illustrates the distribution of various cardiac imaging and radiology reports used in NLP research. Echocardiogram reports, which provide detailed ultrasound-based assessments of cardiac structure and function, dominate the distribution at 51.1\% (n=46). Among the imaging reports, Cardiac MRI reports, which offer high-resolution imaging of cardiac tissues and blood flow, and Chest X-ray reports, which provide radiographic visualization of the heart and lungs, each account for 8.9\% (n=8) of the studies. The remaining imaging types show lower utilization: Cardiac CT reports, offering detailed cross-sectional imaging of cardiac structures, appear in 5.6\% (n=5) of studies; Coronary Angiography reports, which visualize coronary arteries through contrast imaging, in 4.4\% (n=4); Cardiac Catheterization reports, documenting invasive cardiac procedures, in 2.2\% (n=2); and Myocardial Perfusion reports, which assess blood flow to the heart muscle, in 1.1\% (n=1) of studies. ECG reports, which record the electrical activity of the heart through waveform analysis, represent 17.8\% (n=16) of all studies.


The co-occurrence of cardiac imaging and radiology reports reveals notable patterns in how these report types are used together in research. Echocardiogram reports frequently appear alongside chest X-ray and ECG reports, reflecting common research approaches that combine anatomical imaging with functional assessment. Cardiac MRI and CT reports often co-occur, suggesting their complementary roles as advanced imaging modalities. Coronary angiography reports also frequently appear with echocardiogram, ECG, and cardiac catheterization reports, highlighting the integrated, multimodal strategies commonly employed in cardiovascular research.

\subsection{Task Types}

\subsubsection{Analysis based on Categorization}

\begin{sidewaystable}
\caption{Main NLP tasks and example works in Cardiology.}\label{tab1}
\scriptsize
\begin{tabular*}{\textwidth}{@{\extracolsep\fill}p{3cm}p{1cm}p{1.5cm}p{1cm}p{8cm}}
\toprule
Tasks & Freq. & NLP Method & Method Freq. & Example Works \\
\midrule
\makecell[l]{Identification and \\ Classification} & 100 & Rule-based & 25 & \makecell[l]{Identification of Atrial Fibrillation (AF) episodes\\ \citet{RN105} (Linguamatics I2E software, \cite{Milward2005})} \\
\cmidrule{3-5}
& & Deep learning & 48 & \makecell[l]{ \citet{RN24} identified Heart Failure with reduced\\ Ejection Fraction (HFrEF) (Longformer, \cite{Beltagy2020Longformer})} \\
\cmidrule{3-5}
& & Machine learning & 27 & \makecell[l]{ \citet{RN2557} classified cardiac diseases \\ (XGBoost, \cite{chen2016xgboost})} \\
\midrule
\makecell[l]{Prediction} & 69 & Deep learning & 34 & \makecell[l]{\citet{RN2231} predicted cardiovascular disease \\ (BiLSTM-CRF with attention, \cite{Huang2015Bidirectional})} \\
\cmidrule{3-5}
& & Machine learning & 26 & \makecell[l]{\citet{RN120} predicted worsening Heart Failure events \\ (Boosted decision trees, \cite{roe2005boosted})} \\
\cmidrule{3-5}
& & Rule-based & 9 & \makecell[l]{\citet{RN144} predicted cardiac arrhythmia \\ (MYTH algorithm, \cite{Yu2018NILE})} \\
\midrule
\makecell[l]{Information \\ Extraction} & 19 & Rule-based & 7 & \makecell[l]{\citet{RN84} extracted ECG measurements \\ (Custom NLP algorithm, \cite{Torii2011MedTagger})} \\
\cmidrule{3-5}
& & Machine learning & 6 & \makecell[l]{\citet{RN88} performed entity recognition from \\cardiology records (Policy-based Active Learning, \cite{Thompson2019Active})} \\
\cmidrule{3-5}
& & Deep learning & 6 & \makecell[l]{\citet{RN200} extracted cardiovascular disease information \\ (Transformer + LLM, \cite{Devlin2019BERT})} \\
\midrule
\makecell[l]{Automation and \\ Model Evaluation} & 33 & Deep learning & 20 & \makecell[l]{\citet{RN4} performed heart failure hospitalization\\ adjudication. \citet{RN160} evaluated LLMs for cardiooncology.} \\
\cmidrule{3-5}
& & \makecell[l]{Machine \\ learning} & 7 & \makecell[l]{Interactive medical education system~\citep{RN116}\\(Dialogflow API \cite{Google2019Dialogflow}). \cite{RN80} evaluated clinical\\ case management.} \\
\cmidrule{3-5}
& & \makecell[l]{Rule-\\based} & 6 & \makecell[l]{Clinical trial screening~\cite{RN196} (Boolean\\ retrieval system \cite{RN196})} \\
\midrule
\makecell[l]{Text-Guided \\ Generation} & 37 & Deep learning & 37 & \makecell[l]{LLM-based text generation~\citep{chao2025evaluating,sowa2025fine}.\\ Text-to-ECG synthesis~\citep{RN197}.} \\
\bottomrule
\end{tabular*}
\end{sidewaystable}

Tables \ref{tab1} provide an overview of different NLP tasks for cardiology, with detailed information regarding the frequency of all tasks and the distribution of NLP techniques used to approach each task. The results are also accompanied by representative works for each combination of NLP task and method.
The classification schema was adapted from Turchioe et al. \cite{turchioe2022systematic}, with modifications based on the reviewed papers. The top-level categories include information extraction, identification and classification, prediction, automation andmodel evaluation, and text-guided generation.

Among the 258 reviewed papers, identification and classification tasks constitute the largest proportion (38.7\%), with disease cases being the predominant focus. Prediction tasks form the second largest category (26.7\%), particularly centred on disease onset prediction. Information extraction (7.4\%), automationand model evaluation (12.8\%) represent significant areas of application, while text-guided generation (14.3\%) emerges as a novel but fast-developing direction in the field.

\paragraph{Identification and Classification}
The category of Identification and Classification consists of tasks that involve recognizing, categorizing, or labelling clinical entities, documents, or patients based on criteria or patterns. This often involves the interpretation or analysis of extracted information. The application of NLP for identification and classification has become the cornerstone of cardiovascular applications, encompassing several key areas. Disease case identification (40.9\% of identification tasks) represents the primary focus, where NLP models analyze clinical texts to detect various cardiovascular conditions. For example, Nargesi et al. \cite{RN24} developed deep learning models based on the Longformer model architecture \citep{Beltagy2020Longformer} to detect patients with heart failure with reduced ejection fraction (HFrEF). Cardiac measurement extraction (13.6\%) and risk factor identification (11.7\%) form the next tier of frequently addressed tasks, demonstrating the importance of quantitative and risk-related information extraction from clinical documents. \citet{albashayreh2024innovating} innovated the priority classification for older adults with heart failure by re-training a BERT-based classifier \citep{Devlin2019BERT}, significantly improving the accuracy and efficiency of the care priorities identification process.

\paragraph{Prediction}
Prediction primarily includes tasks aimed at forecasting future events, outcomes, or risks based on extracted information and other data. Prediction tasks represent a significant application area in cardiology NLP, with disease onset prediction being the predominant focus (44.0\% of prediction tasks). These applications combine various NLP techniques with machine learning approaches to forecast clinical outcomes (22.0\%) and hospital admissions (20.0\%). For example, \citet{RN2231} employed a sophisticated deep learning approach, using BiLSTM-CRF with attention mechanism, to predict cardiovascular disease onset. Mortality prediction (8.0\%) and disease progression tracking (4.0\%) represent emerging areas within this domain. \citet{chen2025large} proposed a dual attention-network to perform efficient heart failure risk prediction, and leveraged LLMs to process and enrich information provided by the ECG reports.

\paragraph{Information Extraction}
Information Extraction mainly refers to tasks that involve pulling specific pieces of information from unstructured text without making judgments about that information. Information extraction applications focus on transforming unstructured clinical texts into structured data, with attribute and entity extraction being equally prominent (31.6\% each extraction task). These tasks range from extracting specific measurements to identifying broader concepts related to cardiovascular health. An example is \citet{RN84}, which describes a custom, rule-based NLP algorithm for extracting electrocardiogram measurements. Event extraction, relation identification, and entity-relation mapping (each around 10.5\%) represent more complex extraction tasks that capture clinical information's relationships and temporal aspects. To construct a reliable knowledge graph for the heart failure domain, \citet{xu2024knowledge} designed prompting strategies for LLMs to facilitate information extraction from medical guidelines, expert consensus, and professional papers.

\paragraph{Automation and Model Evaluation}
Automation refers to tasks designed to automate clinical processes, provide decision support for healthcare professionals, or enhance efficiency in healthcare delivery. Model evaluation and benchmarking mainly consist of tasks focused on assessing the performance, reliability, and clinical utility of NLP models (especially LLMs) in cardiovascular applications. This includes evaluating communication capabilities, validating domain-specific knowledge, and ensuring safety in clinical settings. 

Automation applications emphasize quality measures (41.2\% of automation tasks) and clinical trial screening (23.5\%), while model evaluation and benchmarking focus on clinical communication evaluation (33.3\%) and domain-specific validation (26.7\%). The emergence of large language models has particularly influenced these areas, as demonstrated by \citet{RN160,RN4-k,shahid2025evaluating}, which evaluated various LLMs in answering cardio-oncology patient queries.

\paragraph{Text-Guided Generation}
Text-guided generation involves the creation of synthetic data, signals, or images based on textual descriptions or clinical narratives in cardiovascular contexts. This category represents an emerging area where text inputs guide the generation of various modalities of medical data, especially with recent fast development of LLM-based techniques. While representing a smaller portion of current applications, text-guided generation demonstrates the field's evolution toward more sophisticated NLP applications. This category shows an equal distribution across text generation, signal generation, and generation evaluation, indicating early exploration of various generative applications in cardiology. Examples of such applications include LLM-based text generation, such as echocardiography reports generation\citep{chao2025evaluating,sowa2025fine}, medical science writing \citep{bhattaru2024revolutionizing}, and ECG report generation \citep{ashfaq2025enhancing}. LLMs are also explored as AI assistants for cardiovascular experts and patients, with applications in scenarios such as Cardiology education \citep{ahmed2025echoing}, consultation for atrial fibrillation and other cardiovascular symptoms\citep{RN157,quer2024potential}, cardiovascular drug development~\citep{ronquillo2024practical}, and clinical quality measurement \citep{adejumo2025fully}, etc. In addition, LLMs are widely utilized to deal with tasks in multiple languages due to their strong multi-lingual NLP capabilities, including Chinese~\citep{ji2025large}, Portuguese~\citep{bruneti2025performance}, and Spanish~\citep{delaunay2024evaluating}, significantly enhancing AI applications for cardiology in low-resource languages.

To summarize, the application of NLP in cardiology clearly emphasises identification, classification, and prediction tasks, particularly in disease detection and risk assessment. With the fast development of modern LLM technologies, emerging areas such as text-guided generation and automated model evaluation suggest future directions for innovation in cardiovascular care.

\subsubsection{Trends in Task Categorization}

\begin{figure}[htbp]
  \centering
  \includegraphics[width=11cm]{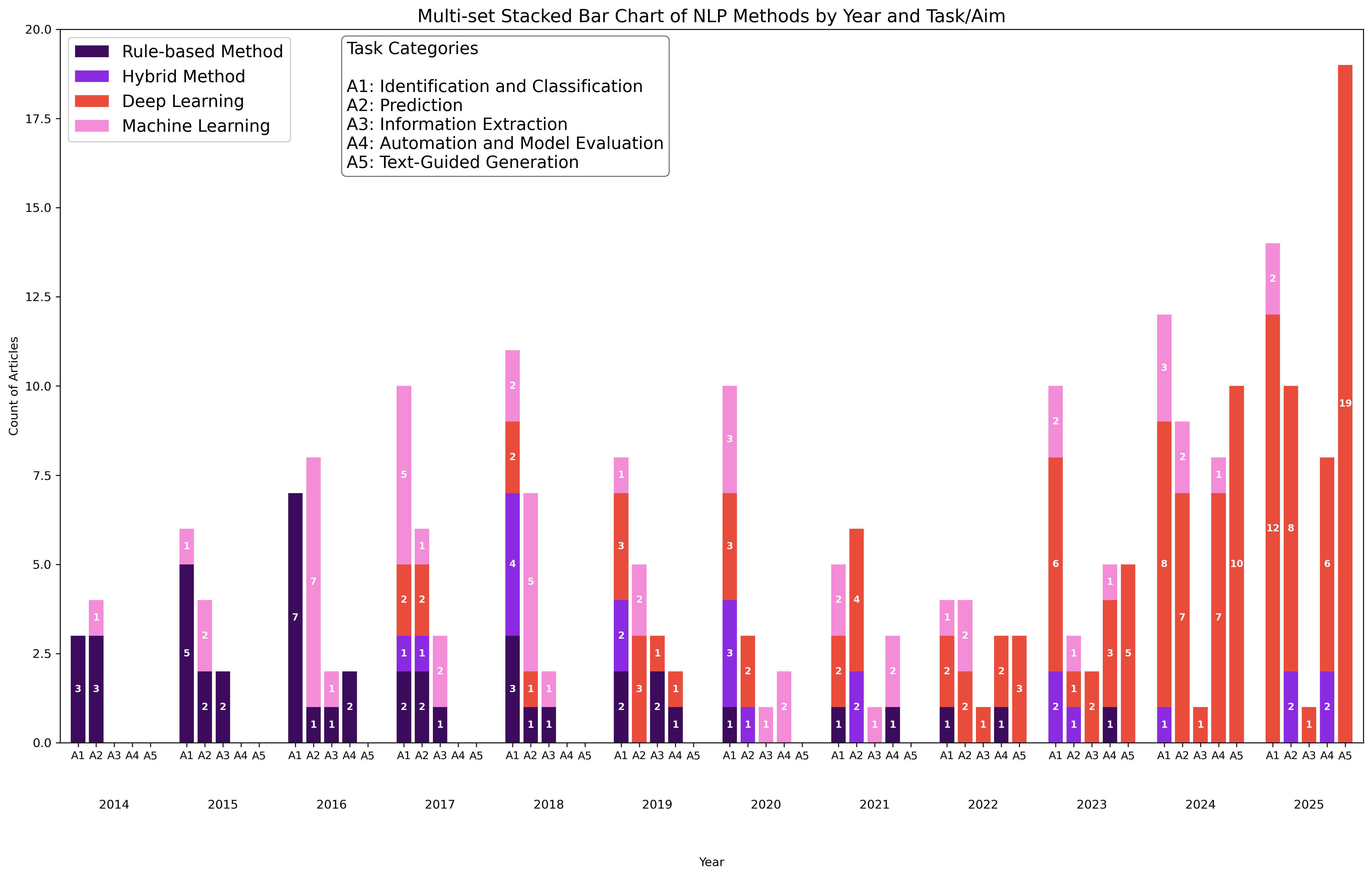}
  \caption{Multi-set stacked bar chart of NLP methods by year and task types.}
  \label{fig:Multiset_stacked}
\end{figure}

Figure \ref{fig:Multiset_stacked} provides a visual depiction of the evolution and changing trends in NLP approaches in cardiology tasks over the past decade. The studies are further split according to the paradigm of their NLP methods. From 2014 onward, we observe an increasing number of studies on all five of the most frequent task types introduced in the previous section. 

Specifically, the task category A1 (Identification and Classification) has been among the most studied fields throughout 2014-2025, showing wide research interest it attracts. Rule-based paradigms were dominant in these tasks from 2024 to 2016, occupying 15 out of 16 studies. Learning-based methods started to perform well in 2017, with a gradual shift from machine learning-based methods to deep learning methods, which have dominated these tasks since 2023.
Studies of category A2 (Prediction) show a steady presence with 6-10 studies per year. Machine learning methods show consistent utilization, but are also being replaced by deep learning-based methods.
Research on A3 (Information Extraction) has maintained a less frequent presence throughout the period and received increased attention during 2017-2019, when machine learning methods were popular and required more structured features. Their attention dropped with the wide application of deep learning methods, especially LLM-based methods, which perform tasks end-to-end and do not require feature extraction.
Research on A4 (Automation and model evaluation) started in 2016 and received fast-growing attention since 2021, with the development of large-scale deep learning and language models, which require fast, accurate, and efficient evaluation of large models such as LLMs. Its development is also benefited by the development in LLMs, such as LLM-as-judge techniques.
The newest category A5 (Text-Guided Generation) is an emerging field since 2022 and has soon received the highest attention in the NLP for cardiology community, with the development of modern deep learning methods, especially LLMs. The field mostly relies on deep learning and LLM techniques, and is becoming the most promising direction for application in practical scenarios with their high-quality AI-generated content for cardiology.

\subsection{Target Cardiovascular Diseases}

\begin{figure}[htbp]
  \centering
  \includegraphics[width=10cm]{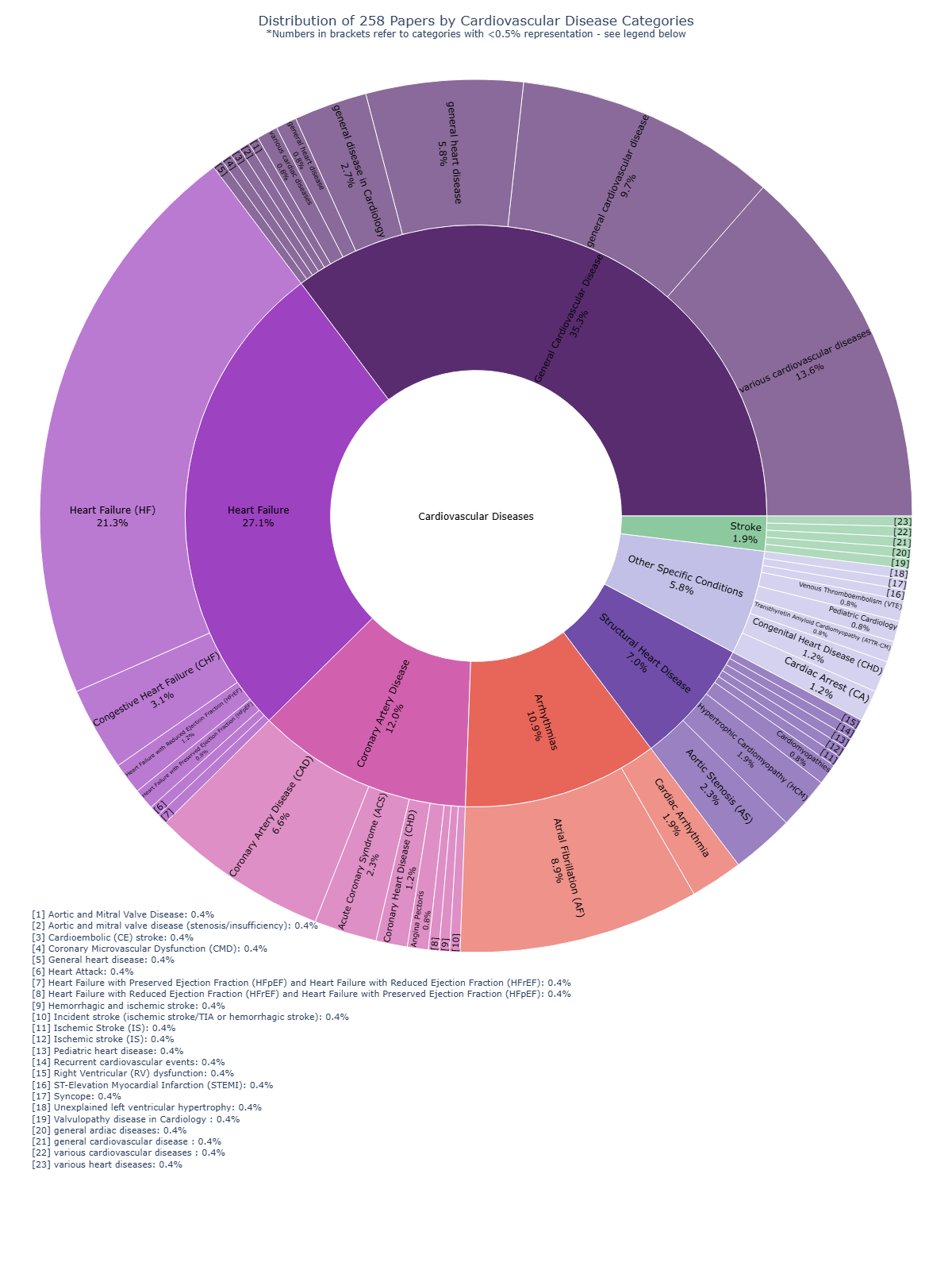}
  \caption{\textbf{Proportions of various types of disease in cardiology addressed in NLP research.} The sunburst chart depicts the percentages of different cardiovascular disease types and subcategories based on the number of reviewed papers that cover them. Especially, the term “Various cardiovascular diseases” is used to characterize works that focus on multiple cardiovascular diseases, while “general cardiovascular disease” characterizes studies that aim to cover all types of cardiovascular diseases without specifying a target category.}
  \label{fig:sunburst_chart}
\end{figure}

Cardiovascular diseases encompass a wide range of conditions affecting the heart and blood vessels. Figure \ref{fig:sunburst_chart} illustrates the range of cardiovascular diseases that are handled using Natural Language Processing (NLP) techniques in the articles that we have reviewed. The classification was performed by manually examining each article to identify the condition(s). 

General Cardiovascular Disease represents the most significant 35.3\% of the studies, encompassing various cardiac conditions and general heart disease research.
Heart Failure accounts for the proportions at 27.1\%, with several specific subtypes included, such as Congestive Heart Failure (CHF) heart Failure with Preserved/Reduced Ejection Fraction (HFpEF/HFrEF), etc. Coronary Artery Disease (CAD) accounts for 12.0\% of studies, including specific conditions like Acute Coronary Syndrome (ACS) and Coronary Heart Disease (CHD). Arrhythmias represent 10.9\% of the research, with Atrial Fibrillation (AF) being the predominant focus. Structural Heart Disease makes up 7.0\% of studies, covering conditions such as Aortic Stenosis (AS) and Hypertrophic Cardiomyopathy (HCM). Other Specific Conditions (5.8\%) include diverse conditions like Cardiac Arrest (CA), Congenital Heart Disease (CHD), and Venous Thromboembolism (VTE). Stroke studies comprise 1.9\% of the research, including both ischemic and hemorrhagic variants. This comprehensive coverage across the cardiovascular disease spectrum demonstrates NLP's versatility in addressing both common and rare cardiac conditions, supporting various aspects of cardiac care from diagnosis to treatment planning. 

For General Cardiovascular Disease, NLP applications are notably diverse, ranging from risk factor identification \cite{RN2247} to automated processing of cardiac imaging reports \cite{RN2214}. A significant portion of these studies focuses on developing generalizable tools for processing various types of cardiac documentation, including efforts to standardize reporting terminology~\citep{martyn2025evaluating,builoff2025evaluating} and extract structured data from unstructured clinical narratives~\citep{mackay2025automated,arnold2025performance}.

Heart Failure is a complex clinical syndrome characterized by the heart's inability to pump blood effectively \citep{Ponikowski2016}. Research on Heart Failure (27. 1\%) focuses mainly on three key areas: early detection and diagnosis using clinical notes and EHR data~\citep{chen2025large,RN24}, prediction of hospital readmissions \cite{RN167,albashayreh2024innovating,shao2025mining}, and identification of worsening heart failure events \cite{RN2215}. For example, some works have used LLMs or other NLP-based methods to extract ejection fraction values from echocardiography reports for heart failure classification and monitoring \cite{RN2548,chao2025evaluating,brown2024race}.

In Coronary Artery Disease research, NLP has been particularly valuable in analyzing cardiac catheterization reports \cite{RN99,RN2301,RN193} and identifying the risk of acute coronary events from clinical documentation~\citep{gao2019automated,RN101,RN213}. Studies in this category have also focused on developing predictive models for disease onset and outcome assessment \cite{RN2289}.

Arrhythmias, which are disorders of the heart's electrical system, are represented significantly in the chart. Atrial Fibrillation (AF), the most common sustained arrhythmia, accounts for a substantial portion of this category. Many studies have concentrated on developing tools for longitudinal monitoring~\citep{RN2501,brennan2025large,Kirchhof2016,feng2025engineering} and identifying treatment patterns~\citep{somani2025understanding,li2025evaluating} in atrial fibrillation management. Other research has also emphasized the automated detection of cardiac rhythm disorders from clinical notes \cite{RN2349}.

Structural Heart Disease mainly consists of conditions affecting the heart's structure, such as valvular diseases and cardiomyopathies. Aortic Stenosis (AS) and Hypertrophic Cardiomyopathy (HCM) are notable subtypes in this category, each requiring specific diagnostic and treatment approaches, which primarily utilize NLP for processing imaging reports and identifying specific structural abnormalities. These efforts have been particularly focused on conditions, such as hypertrophic cardiomyopathy \cite{RN2456,RN2484,RN2438} and aortic stenosis \cite{RN2391,RN91,RN17}.

The above analysis also shows that while certain tasks (e.g., clinical measurement extraction, risk factor identification) are common across disease categories, some other tasks are more prevalent in specific disease categories. For example, readmission prediction is more widely studied in heart failure, while analysis of imaging reports is more frequent in structural heart disease research.

\subsection{NLP Methods for Cardiology}



Existing natural language processing methods in cardiovascular applications can be mostly classified into rule-based, traditional machine learning-based, and deep learning-based methods, where deep learning-based paradigms, especially Large Language Models (LLMs), dominate state-of-the-art solutions. There are also hybrid approaches that leverage more than one of these techniques. We separately introduce these methods as follows.

\subsubsection{Rule-Based Methods}
Rule-based methods use predefined patterns, dictionaries, and expert-crafted rules to extract information from text. These methods have proven particularly effective for extracting standardized measurements and well-defined clinical entities. For example, \citet{RN2374} extracted ejection fraction values from echocardiogram reports, which achieved high precision due to the consistent formatting of these measurements. \citet{RN2215} utilized rule-based approaches to leverage standardized clinical criteria for identifying worsening heart failure events. \citet{olivella2024automatic} used rule-based NLP approaches to filter cases of heart failure patients. \citet{mefford2024rule} used rule-based NLP methods to examine variation in heart failure hospitalizations, considering multiple factors. There are also other works using rule-based statistics (e.g., Bag-of-Words, Skip-gram) as features for machine learning algorithms, successfully applied in scenarios such as random forest \citep{RN2269,RN2285} and support vector machine \citep{RN2512,RN2247,RN2388}.

\begin{table}[htbp]
\caption{A summary of traditional machine Learning and Hybrid Methods for Natural Language Processing in Cardiology.}
\label{tab:ml-hybrid-nlp}
\setlength{\tabcolsep}{4pt}
\begin{tabular*}{\textwidth}{@{\extracolsep{\fill}}p{2.5cm}p{4cm}p{6cm}}
\toprule
Type & Method & Description \\
\midrule
Machine Learning & Random Forest \citep{breiman2001random,RN2438,RN2269,RN99,RN137,RN148,RN52,RN120} & Ensemble learning methods that construct multiple decision trees that combine outputs of individual trees to alleviate overfitting.\\
 & XGBoost \citep{chen2016xgboost,RN2279,RN2557,kamihara2024exploratory} & A scalable tree boosting system that uses a more regularized model formalization to control overfitting. \\
 & Support Vector Machine \citep{cortes1995support,RN195,RN2517,RN67,RN46,RN2416} & A supervised learning model that finds a hyperplane in a high-dimensional space to classify data points. \\
 & Logistic Regression \citep{cox1958regression,RN2274,RN153} & A statistical method for predicting binary outcomes. Simple, fast, and provides easily interpretable results. \\
 & Conditional Random Fields \citep{lafferty2001conditional,RN2560,RN28} & A statistical modeling method often used for structured prediction. Particularly effective for sequence labelling tasks. \\
 & Gradient Boosting \citep{friedman2001greedy,RN51,RN155,RN101} & An ensemble technique that builds new models to correct the errors made by existing models. Known for high performance and flexibility. \\
 & kNN Clustering \citep{cover1967nearest,RN167,RN185,cosin2024safety} & A simple, instance-based learning algorithm that stores all available cases and classifies new cases based on similarity measures. \\
\midrule
Hybrid (with rule-based methods) & Support Vector Machine \citep{RN2512,RN2247,RN2388} & Rule-based method (e.g., dictionary and pattern matching) for feature extraction, and SVM as classifier. \\
 & Conditional Random Fields \citep{RN2318,RN28} & Conditional Random Fields for risk factor identification, Rules or HeidelTime for temporal expression extraction. \\
 & Random forest \citep{RN2269,RN2285} & Regular expression-based feature extraction and Random forest classifier. \\
\midrule
Hybrid (with deep learning) & Neural Network Feature Extractor \citep{RN71,RN58,RN2479,LIU2025114102} & Use neural networks such as RNNs and language models as feature extractors for machine learning methods. \\
\bottomrule
\end{tabular*}
\end{table}

\subsubsection{Traditional Machine Learning-based Methods}
Machine learning methods play a crucial role in NLP research for cardiology, which typically follows a pipeline of data preprocessing, feature extraction, problem modeling, model optimization, and final evaluation. The descriptions and typical examples of the traditional machine learning-based method are illustrated in Table \ref{tab:ml-hybrid-nlp}. A common method is the Support Vector Machine (SVM), a powerful classification algorithm that maps input data to a high-dimensional feature space, and finds the optimal hyperplane that can maximize the margin between samples from different classes \citep{cortes1995support}. SVMs have demonstrated strong performance in processing clinical documentations, including analyzing cardiac catheterization reports \citep{RN99,RN2424}, online clinical discussions~\citep{chandupatla2025scraping}, heart disease classification~\citep{RN195}, and medication use analysis in atrial fibrillation \cite{RN2501}.

Another significant approach is Random Forest, an ensemble learning approach that constructs multiple decision trees and outputs the class that is the mode of the classes (for classification) or mean prediction (for regression) of the individual trees \citep{breiman2001random}. Random forest have demonstrated utility in analyzing cardiac MRI reports for hypertrophic cardiomyopathy detection \citep{RN2438,RN99}, cardiac catheterization
reports~\citep{RN2269}, and echocardiography report narratives~\citep{RN52,RN120}. It is also successfully applied to stroke classification \citep{RN137}.

Building on tree-based machine learning methods, XGBoost (Extreme Gradient Boosting) has emerged as a scalable tree-boosting system, known for its speed and performance in cardiology NLP tasks \citep{chen2016xgboost}. XGBoost has shown effectiveness in temporal analysis, particularly in post-stroke complication prediction \cite{RN2279} and cardiovascular disease detection \cite{RN2557,kamihara2024exploratory}.

Besides the above widely used methods, there are also other effective machine learning algorithms, such as logistic regression \citep{cox1958regression,RN2274,RN153}, used in report categorization and prediction tasks; Conditional Random Fields (CRF) \citep{lafferty2001conditional,RN2560,RN28}, mainly used for information extraction; k-Nearst Neightbor (kNN) Clustering \citep{cover1967nearest,RN167,RN185,cosin2024safety}, proven effective in statistical analysis and risk prediction scenarios. As cardiovascular NLP evolves, researchers have also used hybrid approaches that leverage the strengths of multiple techniques. One common hybrid approach combines rule-based systems with machine learning classifiers \citep{RN2285}. The combination of rule-based and machine learning approaches for cardiovascular risk factor identification \citep{RN2432,duminuco2024development} also achieved higher accuracy by integrating domain expertise with statistical learning. In heart failure subtype classification \citep{RN2217}, hybrid methods effectively combined standardized diagnostic criteria with pattern recognition. For complex tasks like analyzing cardiac imaging reports \citep{RN2282,east2025enhancing}, hybrid frameworks successfully integrated multiple analytical approaches to improve performance.

\begin{table}[htbp]
\caption{A summary of deep learning methods for Natural Language Processing in Cardiology.}
\label{tab:deep-learning-nlp}
\tiny
\setlength{\tabcolsep}{4pt}
\begin{tabular*}{\textwidth}{@{\extracolsep{\fill}}p{2.5cm}p{4cm}p{6cm}}
\toprule
Type & Method & Description\\
\midrule
Neural Networks & Convolutional Neural Network \citep{LeCun1998,RN164,RN2377,RN32} & CNN mainly consists of convolutional layers, pooling layers, and FFN layers, which aims to capture dependencies among near features. \\
 & Recurrent Neural Networks: LSTM \citep{Hochreiter1997,RN134,RN147}, GRU \citep{Cho2014,RN2288}, BiLSTM \citep{bahdanau2014neural,RN2283,RN2231} & A family of neural networks that performs inferences on sequential data, designed to capture time-series dependencies. \\
 & Hybrid Methods~\citep{RN78,RN2433,RN111} & Methods that combine multiple model, architectures, such as CNN-LSTM, CNN-MLP, and LSTM-BERT, leveraging all their advantages. \\
\midrule
Pre-trained Language Models & BERT \citep{Devlin2019,RN2424,RN2284,RN2555,RN2281,RN175,RN188}, RoBERTa \citep{Liu2019,RN2514} & Transformer-based language model families, pretrained on large-scale general data in an auto-encoder manner. They are normally used as feature extractors. \\
 & BioBERT \citep{Lee2020,RN2424,RN2369,RN2555}, ClinicalBERT \citep{Alsentzer2019,RN57,adejumo2024natural} & Transformer-based langauge model families, pre-trained on large-scale biomedical corpora, making it particularly suited for biomedical text mining tasks. \\
\midrule
Large Language Models & LLMs for Prediction \citep{li2025ai,chen2025large,mahmoudi2025comparative,ji2025large,duminuco2024development,adatya2025enhancing} & Large Language Models (LLMs) are based on the Transformer architecture and pre-trained on massive-scale web data in an auto-regressive manner. It can be used to perform prediction tasks for future risks in a prompt-based manner. \\
 & LLMs for Text Generation \citep{RN2363,RN160,RN192,RN2534,RN157} & LLMs can be used to generate useful texts for cardiology, such as CT reports, answers to related questions, and knowledge sources. \\
& LLMs as AI Assistants \citep{RN117,kozaily2024accuracy,el2024accuracy,singh2025enhancing,RN80,RN2384,RN2297,tran2025role} & LLMs can be used as AI assistants to provide clinical advice for various cardiology diseases for patients. It can also help write cardiology reports in multiple scenarios for doctors.  \\
 & LLMs for Information Extraction \citep{RN2384,RN2297,xu2024knowledge,shang2025evaluating,chi2025echollm,tully2022classification} & LLMs can be used to process natural language-based text and extract structured or unstructured information (e.g., relation extraction, classification, knowledge graph, structured reports) to facilitate the diagnosis process. \\
& Retrieval-augmented Generation~\citep{adejumo2025fully,hayama2025performance,lakhdhar2025chatcvd,parameswaran2025evaluating} & Based on LLMs, retrieval-augmented generation boosts the factuality and richness of their generated text by introducing information extracted from external structured or unstructured knowledge sources.\\
\bottomrule
\end{tabular*}
\end{table}

\subsubsection{Deep Learning-based Methods} 
Deep learning methods aim to perform complex pattern recognition and feature extraction from large-scale raw data through deep neural network training, which has significantly improved the performance of NLP methods in cardiology, even with a paradigm shift. The descriptions and typical examples of deep learning-based methods are presented in Table \ref{tab:deep-learning-nlp}. The related methods can be categorized into neural network-based methods, pre-trained language model-based methods, and modern Large Language Model (LLM)-based methods. 

\paragraph{Neural Networks}
The early neural network-based methods use different customized architectures based on the features of the target tasks and train a small to moderate-scale model on in-distribution data from scratch. Common architectures include include Convolutional Neural Networks (CNNs), which mainly consist of convolution layers and pooling layers. CNNs excel at extracting local features from text thanks to the convolution layers, which have been applied to tasks such as analyzing thoracoabdominal CT reports \citep{RN2377}, heart failure prediction from clinical notes~\citep{RN164,marti2025natural}, and analysis on the patterns of atrial fibrillation~\citep{RN32}. Recurrent Neural Networks (RNNs) is another widely used model family, mainly consisting of Long Short-Term Memory (LSTM) and Gated Recurrent Unit (GRU) networks, which are designed to capture dependencies sequentially in long-range texts. These architectures have been proven effective in analyzing Cardiovascular risk from sequential data~\citep{RN134} and information extraction from clinical data~\citep{RN147,RN2288}. Advanced RNN architectures such as Bi-LSTM are also widely applied in cardiovascular disease/events prediction~\citep{RN2231,RN2283}. Hybrid methods combine different model architectures to achieve superior performance over a single model. Successful combinations include CNN-LSTM, CNN-MLP, and LSTM-BERT models, achieving outstanding performance in scenarios including heart failure prognosis~\citep{RN78} and processing of electrocardiogram~\citep{RN2433} and radiology reports~\citep{RN111}.

\paragraph{Pre-trained Language Models}
Pre-trained Language models (PLMs) aim to learn general knowledge via pre-training on large-scale general web text data (usually several billion tokens). The models are usually based on the Transformer architecture and further adapted to the target task by further fine-tuning on a small amount of domain-specific data. From 2019 onwards, the BERT model family has become a dominating paradigm in the cardiology field, successfully applied in tasks such as cardiac detection and prediction~\citep{RN2555,RN2281}, information extraction from clinical reports~\citep{RN2284,RN175,RN188}, etc. Other improved model varaints such as RoBERTa are also widely used~\citep{RN2514}. Another line of PLM works introduces task-specific knowledge via pre-training on large-scale domain data. For example, BioBERT is pretrained on biomedical data and outperforms general PLMs on prediction tasks on EHRs~\citep{RN2424}, coronary catheterization reports~\citep{RN2555}, and clinical notes~\citep{RN2369}. ClinicalBERT performs well on information extraction and analysis for heart failure patients~\citep{RN57,adejumo2024natural}.

\paragraph{Large Language Models}
Since the release of ChatGPT in 2022, Large Language Models (LLMs) have marked a major paradigm shift in NLP. LLMs are huge Transformer-based models (usually billions of parameters) pre-trained on massive scale of web data (usually trillions of tokens) to obtain outstanding general problem-solving capabilities and strong in-context learning capabilities, which enable LLMs to adapt to new tasks without modifying the parameters. LLMs have become an emerging dominant paradigm in cardiology, significantly outperforming previous methods in risk prediction for cardiovascular diseases~\citep{li2025ai,ji2025large,duminuco2024development}, heart failure~\citep{adatya2025enhancing}, and automated echocardiography report analysis~\citep{mahmoudi2025comparative}. Many works also explored utilizing LLMs for assisting diagnosis of related diseases, including as knowledge sources~\citep{qiu2025cvdllm}, agent for symptom description and management~\citep{tian2025novel}, and interpreters of existing reports~\citep{bozyel2025large}. LLMs also show generalized ability to handle signals from multiple sources, including ECG signals~\citep{chen2025large,yang2025ecg,lee2025comparative}, text-based cardiology disease reports~\citep{bazoge2025assessing,10.1093/jamia/ocae085,wehbe2025charting}, and educational materials~\citep{bhattaru2024revolutionizing,builoff2025evaluating,rullo2025interdisciplinary,hijazi2025priming}. Due to their auto-regressive nature, LLMs are also widely used in text generation tasks, successfully applied in question answering for cardiology-related topics~\citep{RN2363,RN160,RN192,RN2534}, AI assistants for clinical advice on myocardial infarction~\citep{RN117,kozaily2024accuracy}, atrial fibrillation~\citep{RN157,el2024accuracy}, cardiovascular rehabilitation~\citep{singh2025enhancing}, and cardiac electrophysiology procedures~\citep{sritharan2025utility,ahmed2025primer}, and writing summaries of cardiology reports~\citep{RN80}, cardiovascular magnetic resonance reports~\citep{RN2384,wang2025automated}, and education materials~\citep{RN2297}, etc. LLMs are also explored as research and decision-support tools for domain experts~\citep{tran2025role,wang2025current,jung2025clinical}. In practical scenarios, the high-quality LLM-generated content can significantly lower the burden of domain experts and the cost of patients for professional advice. Another application focuses on extracting information from texts via prompt-based instructions. For example, \citet{xu2024knowledge} leveraged LLMs to extract structured knowledge from heart failure-related texts and built the first knowledge graph in this domain. \citet{shang2025evaluating} evaluated the factuality of extracted Cardiovascular-Kidney-Metabolic syndrome knowledge within LLMs to ensure their reliability. \citet{chi2025echollm} utilized LLMs to extract echocardiogram entities in a flexible manner. \citet{tully2022classification} compared LLMs to rank the priorities for surgical patients needing preoperative cardiac evaluations. \citet{vasisht2025automated} converted text-based cardiovascular diagnostic reports into structured datasets with LLMs. Finally, due to LLMs' high hallucination rate in knowledge-intensive scenarios, the retrieval-augmented generation (RAG) technique has been explored in cardiology, which aims to enhance LLM-generated content with retrieved knowledge or texts. For example, \citet{adejumo2025fully} incorporated structured information from inpatient heart failure notes to combine into the context of LLMs. \citet{hayama2025performance} comprehensively evaluated RAG-based LLMs and general LLMs in Cardiology specialist examinations in Japan, proving external knowledge useful for QA tasks. \citet{lakhdhar2025chatcvd} used RAG in developing AI assistants for personalized cardiovascular risk assessment, and \citet{parameswaran2025evaluating} further proved its effectiveness by using RAG to enhance LLM performance in providing nutrition information for cardiovascular diseases.

\begin{figure}[ht]
  \centering
  \includegraphics[width=11cm]{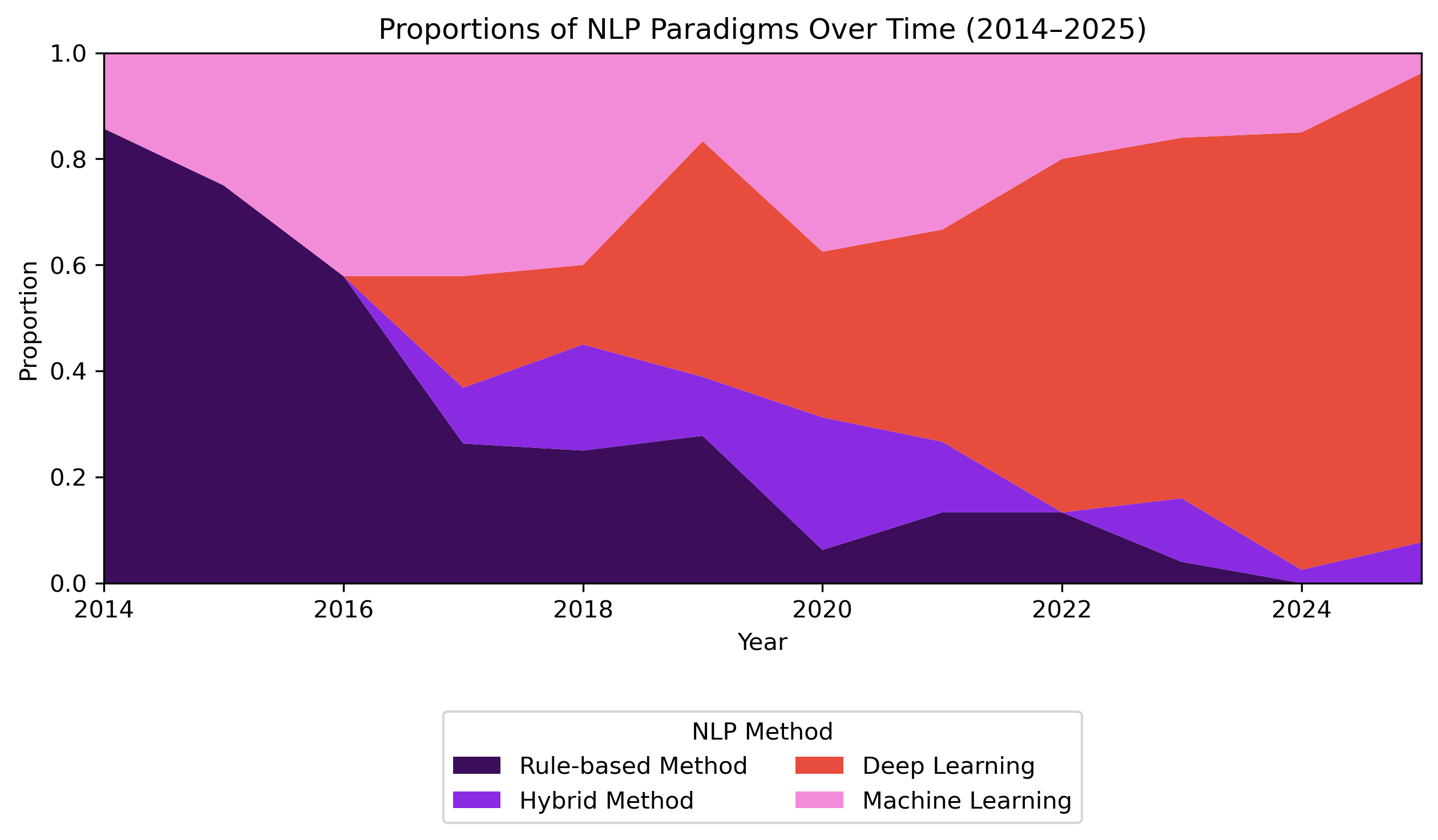}
  \caption{Trends in NLP methods used in Cardiology by year.}
  \label{fig:nlptimestacked}
\end{figure}

\begin{figure}[ht]
  \centering
  \includegraphics[width=11cm]{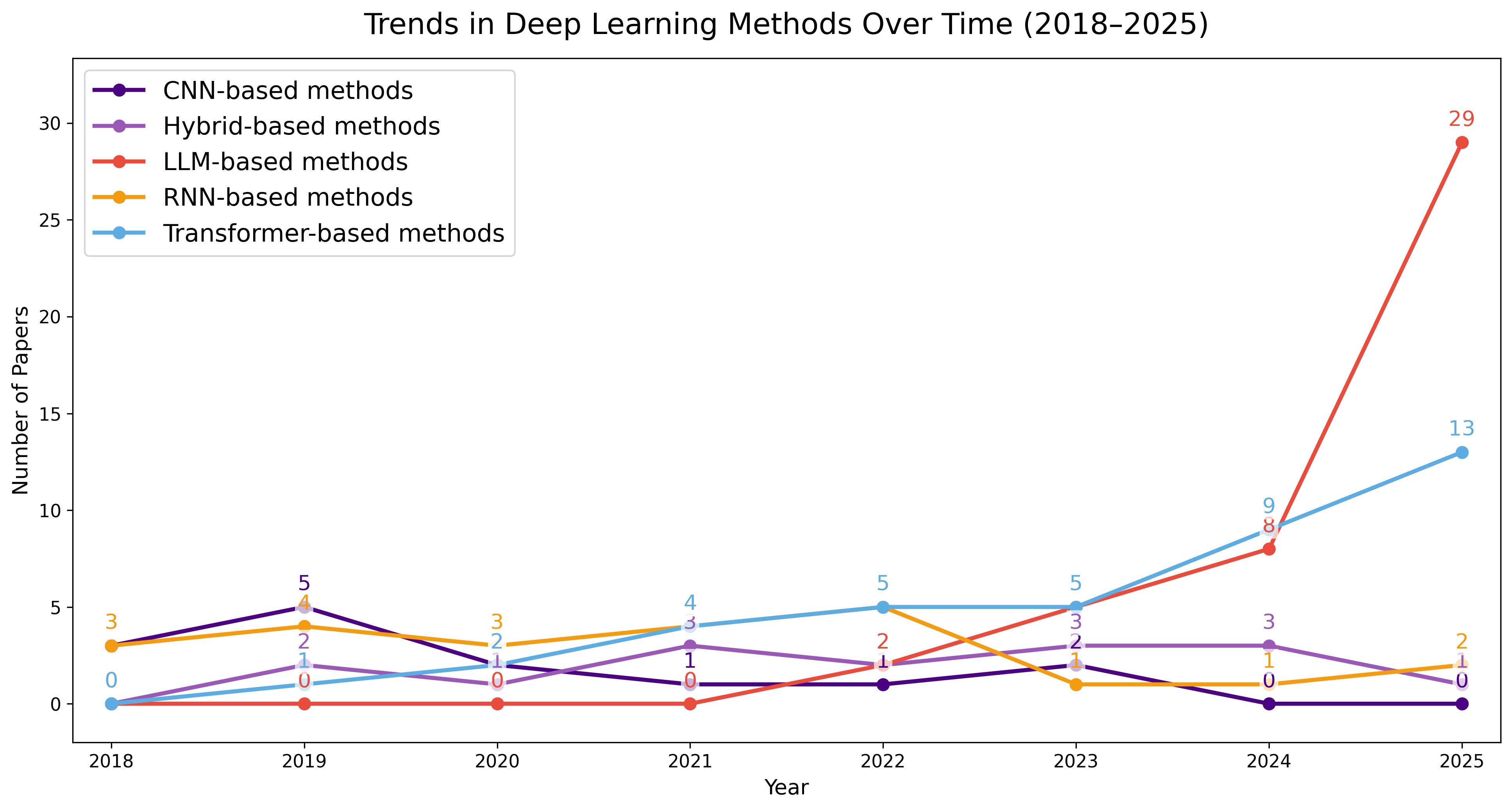}
  \caption{Deep Learning-based NLP methods used in Cardiology by year.}
  \label{fig:deeptimeline}
\end{figure}

\subsubsection{Trends in NLP Methods for Cardiology} 
We further provide an overview of the trends in the above NLP methods along the time dimension. As illustrated in Figure \ref{fig:nlptimestacked}, the distribution of NLP paradigms in cardiology has evolved significantly over time. The proportion of rule-based methods, though dominates in the early stage (2014-2016), has significantly decreased since 2017, mainly because of its significant lack of flexibility and generalization ability across different cases. Traditional machine learning approaches have maintained a consistent presence throughout different stages, as they possess pattern recognition ability across cases and are easy to implement. They thrived between 2018 to 2022, but show a decreasing trend gradually replaced by deep learning-based methods. Deep learning methods started to be applied in cardiology since 2017 and have shown consistent improvement from 2019 onwards, with the fast development of Transformer-based methods and PLMs. They have become the dominant technique to date, mainly due to the emergence of LLMs and their breakthrough in unifying the paradigm across sub-fields in cardiology and outstanding performance. Deep learning methods also benefit from the fast development of modern computational resources and facilities, which facilitate their deployment in practical scenarios, and are expected to remain the mainstream research direction.

To provide a specific view for the trend in deep learning-based methods, Figure \ref{fig:deeptimeline} breaks them down in representative model architectures and paradigms, including CNNs, RNNs, Transformer (PLMs), and LLMs. According to the results, the overall numbers of deep learning-based methods has been fast growing since 2018, showing their dominant trend in cardiology. Specifically, CNN-based methods and RNN-based approaches, though maintaining steadily throughout the period, have been significantly outweighted by other emergent paradigms, such as Transformer-based methods and LLMs. Transformer-based methods have shown significant improvements since 2019, especially PLM-based methods such as BERT \cite{RN2424,RN2284,RN2555,RN2281}, which have been widely adopted for various cardiovascular applications, especially in identification, classification, risk prediction, and information extraction tasks. The year 2022 marks a surge of Large Language Models in cardiology, with capable generative LLMs such as GPT-4 \cite{RN2384,RN117} and ChatGPT \cite{RN2297,RN2363}, which substantially improves model performance in multiple crucial scenarios, including cardiovascular report analysis and first-aid for heart failure. LLMs also mark a new phase in the application of generative AI to cardiology and clinical tasks, significantly improving the quality of generated content in scenarios such as writing of education materials and AI assistants. The emergence of LLMs also encourages the exploration of LLM as knowledge source for machine learning/deep learning methods.
Hybrid methods that combine deep learning techniques with high-quality human supervision show a consistent presence, indicating their continued utility in cardiovascular applications where both domain expertise and advanced pattern recognition capabilities are required.

In summary, these temporal trends indicate a consistent presence of different approaches in parallel rather than new paradigms completely replacing old ones, and method selection in cardiovascular NLP continues to be driven by specific task requirements. However, advanced NLP techniques such as Transformer and LLMs are increasingly dominating related works in recent years, showing their superior performance in more application scenarios related to cardiology, which is consistent with the overall trend in AI developments.

\section{Discussion}\label{sec12}
As the latest review of NLP application to cardiology, we first investigate whether the prospects of previous reviews \citep{turchioe2022systematic} have been revealed with the latest development of NLP techniques. Firstly, application of NLP tools to a broader range of cardiovascular diseases has been achieved (as shown in Figure \ref {fig:sunburst_chart}), with NLP methods covering all major sub-fields in cardiology. Secondly, more sophisticated NLP approaches have been applied to improve performance on complex tasks, especially with the advanced PLM and LLM-based methods. Thirdly, novel methods have been developed to extract information from unstructured data for cardiology, such as primary care notes, patient-generated health data, social media content, and inpatient physician notes, which were once hard to process.

Although promising results have been obtained using
traditional machine learning and deep learning methods, especially LLM-based methods, several challenges remain
for NLP application in cardiology that require further research.
Herein, we introduce some key challenges and future research directions.

\subsection{Main Challenges}

\paragraph{Interpretability} 
Improving the interpretability of deep learning and machine learning models remains an underexplored yet important area of research. From an interpretability perspective, rule-based NLP approaches offer notable advantages due to their inherent transparency and the level of control they provide. Although these systems demand considerable human effort, their continued use in real-world clinical settings can be partly attributed to their ability to let researchers explicitly define criteria of interest, thereby ensuring a clear and auditable decision-making process. In contrast, the “black-box” nature of deep learning models makes it difficult to explain why they produce specific predictions. Some interpretable deep learning methods can highlight the keywords, phrases, or sentences that influence a model’s decision, and the interpretation of LLMs decisions have been improved via chain-of-thought techniques. However, the underlying reasoning linking these textual elements to the final output often remains opaque and unreliable~\citep{korbak2025chain}. This limitation poses particular challenges in the clinical domain, where even correct predictions may occur by coincidence, and the model’s reasoning process may be fundamentally flawed.

\paragraph{Trustworthiness} 
The majority of NLP-based studies in cardiology have been developed to support healthcare and clinical professionals. However, there is also substantial potential for AI-driven tools to empower patients in understanding and managing their cardiac conditions. For such tools to achieve widespread adoption and sustained use, patients must have confidence in their accuracy, reliability, and usefulness. Despite this, few studies have explored patients’ perceptions, understanding, and attitudes toward different AI-based tools. Moreover, challenges surrounding the implementation of NLP algorithms in clinical practice (e.g. costs, clinical workflows, time burden) have still not been in-depth discussed in many papers concerning NLP in Cardiology, which can significantly increase the uncertainty and unexpected cost of deploying these algorithms in practical scenarios.

\paragraph{Data Privacy and Compliance with Regulations}
Existing works in NLP for cardiology mostly rely on EHR or clinical data that contain protected health information. However, ensuring de‐identification or anonymization is often incomplete or difficult. Even when data are purportedly de‐identified, there is risk of re‐identification through linkage attacks or via models that memorize sensitive textual elements~\citep{sousa2023keep}. Existing research also lacks clarity in demonstrating compliance with data privacy rules.
It is also difficult to ensure generative models, such as LLMs, are trained, deployed, and accessed in ways that satisfy minimum necessary data rules and governance~\citep{moell2025harm}.
Together, these issues limit data privacy and transparent supervision for compliance with regulations for use of NLP in cardiology.

\subsection{Future Directions}

\paragraph{Interpretable Large Language Models}
Based on the analysis of research trends provided in this article, LLMs are becoming increasingly dominant across most task categories in cardiology, mainly due to their unified generative paradigm and outstanding generalization capability across tasks. One direction is to develop advanced hybrid models that can combine other interpretable and light-weight methods with LLMs to enhance their performance and efficiency, such as rule-based methods, graph machine learning, and retrieval-augmented generation~\citep{fu2024research,RN92}. 
The interpretability of LLMs has also been investigated and improved \citep{singh2024rethinking} in other fields, which shows great potential for application in NLP for cardiology. Existing studies in other areas have already focused on designing LLMs that can self-explain via chain-of-thought prompting strategies~\citep{wei2022chain,lyu2024language}. Thus, the development of LLMs that can understand and generate in the cardiology field in an interpretable manner constitutes an important future development direction.

\paragraph{Multi-Modal Methods}
Another promising direction is multi-modal learning, which combines evidence from text with other modalities to further enhance model performance in cardiology tasks. Building upon state-of-the-art multi-modal LLMs~\citep{mckinzie2024mm1}, there are emerging multi-modal methods that have shown promising results in integrating information from ECG reports and images~\citep{lee2025comparative}, visual QA from ECG image interpretations~\citep{seki2025assessing}. In future work, there is significant potential to combine other cardiology images, such as cardiac MRI images, with narrative clinical text. Other modalities also bear potential as reliable information sources, including knowledge graphs~\citep{xu2024knowledge,shang2025evaluating}, other ECG signals~\citep{RN197}, and heart rate variation signals~\citep{sassi2015advances}, which can be leveraged to further improve model performance on disease diagnosis and risk prediction tasks. 

\paragraph{Open-source Tools}
Previous open-source NLP tools like the GATE-based NLP system, I2E software, and cTAKES with UMLS CUIs are widely used in tasks like outpatient worsening heart failure identification \citep{RN208} and information extraction from echocardiography reports~\citep{RN65}. These studies proved the usefulness and reliability of open-source NLP tools across health systems and use cases. Further development of LLM-based methods in the cardiology field can also significantly benefit from more powerful open-source LLM series for public use and high-quality open-source training data that corresponds to data privacy and related regulations. More open-source datasets for low-resource languages can also significantly benefit future research on NLP applications in multi-lingual scenarios. Overall, future research on open-source tools bears huge potential in democratizing AI usage in cardiology.

\section{Conclusion}\label{sec13}
This review provided a comprehensive and multi-dimensional overview of NLP-based research for cardiological applications over the past decade. Based on the discussion within this review, we have witnessed a significant shift in the paradigm of NLP methods used in cardiology research, with the emergence of various application scenarios in cardiology and increasingly powerful NLP methods such as LLMs. 
Existing NLP algorithms have been successfully applied in cardiology and domain-specific modifications can further enhance model performance. State-of-the-art LLMs continue to show great potential in both enhancing model performance and improving interpretability, which have become a major paradigm in the field of cardiology. Nevertheless, the common challenges of low interpretability within these systems have not been sufficiently addressed. Improving the interpretability of LLMs and introducing multi-modal supervision signals are promising directions for future work, and a more open-source community is the key for deployment and democratized access of NLP methods for cardiology.

\section*{Acknowledgments}
This work is supported by British Heart Foundation Manchester Centre of Research Excellence (RE/24/130017). This work is supported in part by The University of Manchester President's Doctoral Scholar award.

\bibliography{sn-bibliography}

\end{document}